\documentclass[letterpaper, 10 pt, conference]{ieeeconf}  

\IEEEoverridecommandlockouts                              

\usepackage[export]{adjustbox}            
\usepackage{multirow}
\usepackage{algorithm}                    
\usepackage[noend]{algpseudocode}         
\usepackage{amsmath}                      
\usepackage{amssymb}                      
\usepackage{cite}                         
\usepackage{graphics}                     
\usepackage[usenames,dvipsnames]{xcolor}  
\usepackage{subfigure}

\usepackage{url}
\usepackage{hyperref}

\makeatletter
\newcommand\fs@betterruled{%
  \def\@fs@cfont{\bfseries}\let\@fs@capt\floatc@ruled
  \def\@fs@pre{\vspace*{5pt}\hrule height.8pt depth0pt \kern2pt}%
  \def\@fs@post{\kern2pt\hrule\relax}%
  \def\@fs@mid{\kern2pt\hrule\kern2pt}%
  \let\@fs@iftopcapt\iftrue}
\floatstyle{betterruled}
\restylefloat{algorithm}
\makeatother


\newcommand{\algorithmicinput}{\textbf{Input:}}
\newcommand{\algorithmicoutput}{\textbf{Output:}}
\newcommand{\INPUT}{\item[\algorithmicinput]}
\newcommand{\OUTPUT}{\item[\algorithmicoutput]}
\newcommand{\CALL}[1]{\textsc{#1}}
\newcommand{\drprm}{Dynamic Region Sampling with PRM}

\newcommand{\dofs}{{\sc dof}s}
\newcommand{\cspace}{\ensuremath{\mathcal{C}_{space}}}
\newcommand{\cspaces}{\ensuremath{\mathcal{C}_{spaces}}}
\newcommand{\cfree}{\ensuremath{\mathcal{C}_{free}}}
\newcommand{\cobst}{\ensuremath{\mathcal{C}_{obst}}}

\newcommand{\rrt}{RRT}

\newcommand{\drbrrt}{Dynamic Region-biased RRT}

\newcommand{\qskeleton}{Query Skeleton}

\newcommand{\red}[1]{#1}

\title{\LARGE \bf
Scalable Multi-robot Motion Planning for Congested Environments\\ With Topological Guidance
}

\author{Courtney McBeth$^{1}$, James Motes$^{1}$, Diane Uwacu$^{2}$, Marco Morales$^{1}$, and Nancy M. Amato$^{1}$%
\thanks{$^{1}$Courtney McBeth, James Motes, Marco Morales, and Nancy M. Amato are with the Parasol Lab, Department of Computer Science,
        University of Illinois Urbana-Champaign, Champaign, IL 61820 USA
        {\tt\small \{cmcbeth2, jmotes2, moralesa, namato\}@illinois.edu}}%
\thanks{$^{2}$Diane Uwacu is with the Texas A\&M University Department of           Computer Science and Engineering, College Station, TX 77840 USA
        {\tt\small duwacu@tamu.edu}}%
\thanks{This work was supported in part by Foxconn Interconnect Technology (FIT) and the Center for Networked Intelligent Components and Environments (C-NICE) at UIUC.}
}

\begin{document}

\maketitle
\thispagestyle{plain}
\pagestyle{plain}

\begin{abstract}

Multi-robot motion planning (MRMP) is the problem of finding collision-free paths for a set of robots in a continuous state space.
\red{The difficulty of MRMP increases with the number of robots and is exacerbated in environments with narrow passages that robots must pass through, like warehouse aisles where coordination between robots is required.}
In single-robot settings, topology-guided motion planning methods have shown improved performance in these constricted environments.
\red{In this work, we extend an existing topology-guided single-robot motion planning method to the multi-robot domain to leverage the improved efficiency provided by topological guidance.}
We demonstrate our method's ability to efficiently plan paths in complex environments with many narrow passages, scaling to robot teams of size up to \red{25} times larger than existing methods in this class of problems.
By leveraging knowledge of the topology of the environment, we also find higher-quality solutions than other methods.

\end{abstract}

\section{INTRODUCTION}

Multi-robot systems have become ubiquitous in many \red{settings} such as autonomous factories and warehouses. 
\red{When the motions of two or more robots are in conflict, highly coordinated multi-robot motion planning (MRMP) is required to avoid collisions with each other and with the environment.}

Existing MRMP approaches perform well in open environments but struggle with narrow passages 
(e.g. warehouses aisles)
due to the difficulty of avoiding collisions with obstacles within tight spaces.
With multi-robot teams, narrow passages may be introduced or exacerbated by inter-robot collisions, further complicating planning.
\red{
There are three classes of MRMP approaches: decoupled, which plan a path for each robot separately and offer the least amount of coordination, coupled, which plan in the \textit{composite space}, which incorporates the degrees of freedom of all robots, and hybrid, which offer a mix of coupled and decoupled planning.
Decoupled approaches (e.g., Decoupled PRM~\cite{sl-uppccdpmrs-2002}) allow for linear scaling with the number of robots; however, they lack the coordination necessary to resolve complex inter-robot collisions.
Coupled methods (e.g., Composite RRT~\cite{l-rrtntpp-1998} and Composite PRM~\cite{sl-uppccdpmrs-2002}) provide this coordination but consider a large search space that becomes computationally intractable for large teams.
Hybrid methods (e.g., CBS-MP~\cite{smsa-rmmpucs-21} and M*~\cite{wc-sefmpp-15}) leverage the scalability of decoupled methods along with an increased level of coordination. 
However, in environments with narrow passages where inter-robot collisions are likely, their performance is limited by the time spent resolving conflicts.
In this work, we propose a coupled method that exploits topological guidance to more intelligently and efficiently search the composite space.
}

\begin{figure}
    \centering
    \includegraphics[width=0.48\textwidth]{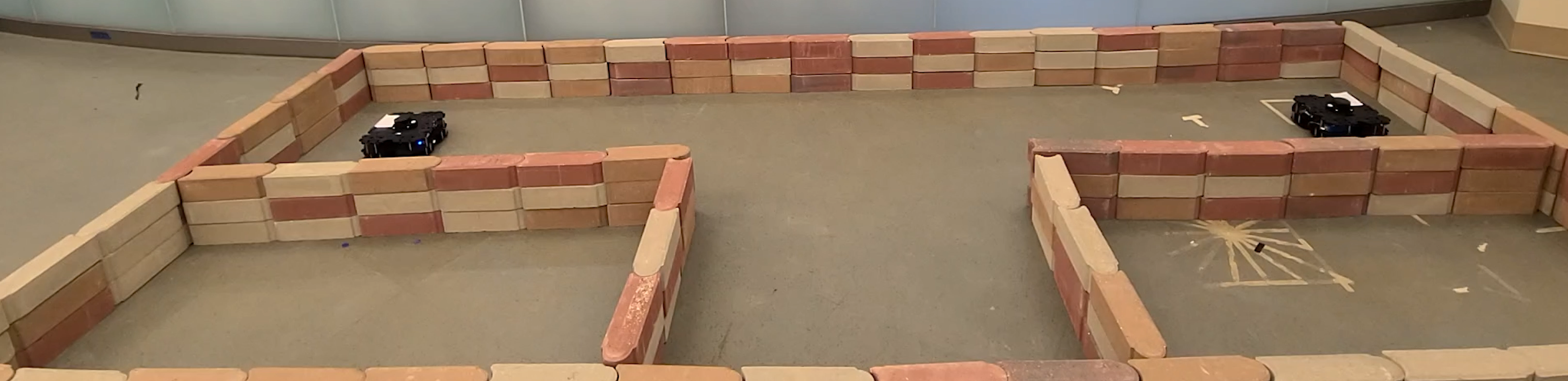}
    \caption{We demonstrate our method on a variety of physical and simulated environments. Video: \red{\url{https://youtu.be/xP7mSWxdYfs}}}
    \label{fig:physical-inlet}
\end{figure}    

Guided planning methods leverage external information to efficiently find paths.
\red{
Topology-guided methods~\cite{dsba-drbrrt-16, suda-tgrcdrs-20, rsb-beaeisbmp-2014} exploit representations of the environment to direct motion planning through narrow passages.
Topological skeletons are embedded graphs encoding the environment. 
Prior work~\cite{dsba-drbrrt-16, suda-tgrcdrs-20} has explored guiding planning around a skeleton.}

\red{
In this work, we extend topological skeleton guidance to multi-robot systems to provide the level of coordination required for large teams in environments with narrow passages.}
We present \textit{Composite Dynamic Region-biased Rapidly-exploring Random Trees} (CDR-RRT), a coupled MRMP approach that leverages knowledge of the workspace topology, leading to improved planning times while retaining probabilistic completeness. We demonstrate significantly improved scalability compared to existing state-of-the-art methods, successfully finding paths for teams of size up to \red{25} times larger in complex environments where inter-robot collisions are likely.
Our contributions include:
\begin{itemize}
    \item The development of \red{novel} composite-space analogs of workspace skeletons and sampling regions.
    \item \red{A scalable, probabilistically complete MRMP method that leverages topological guidance to address problems that require high levels of coordination.}
    \item An experimental validation in a variety of congested and open environments with scaling robot team sizes.
\end{itemize}


\section{PRELIMINARIES AND RELATED WORK}
In this section, we discuss \red{the motion planning problem and research} in both the multi-robot motion planning and guided motion planning domains.

\subsection{Motion Planning Preliminaries}
A robot's \textit{degrees of freedom} (\dofs) include its position and orientation in the \textit{workspace}, the 2D or 3D space within which the robot physically exists, as well as other configurable values such as joint angles. 
A \textit{configuration} is a set of values describing the robot's \dofs. 
The \textit{configuration space} (\cspace) is the set of all robot configurations~\cite{lw-apcfpapo-79}. 
The \textit{free space} (\cfree) is the subset of \cspace{} that only contains valid configurations \red{(e.g. configurations not in collision with obstacles).}
The \textit{obstacle space} (\cobst) contains all configurations that are \red{not valid.}
Given a start configuration, $q_{start}$, and a goal configuration, $q_{goal}$, the \textit{motion planning problem} strives to find a path from $q_{start}$ to $q_{goal}$ through \cfree.

Searching the entire \cspace{} is intractable~\cite{ss-otpmpiigtfctporam-83,c-crmp-88}, resulting in the emergence of sampling-based motion planning algorithms~\cite{kslo-prpp-96,l-rrtntpp-1998}. 
These algorithms forego completeness guarantees in favor of faster planning and \textit{probabilistic completeness}, meaning that the probability of finding a solution, if one exists, converges to 1 in the upper limit of planning time. Unfortunately, these randomized sampling techniques suffer in constrained environments~\cite{hlk-pfprp-06}, \red{where they face the \textit{narrow passage problem}. This refers to the difficulty of sampling valid configurations within narrow corridors that by volume make up a small proportion of the freespace.
}

The underlying sampling-based motion planning algorithm that forms the basis of our method is Rapidly-exploring Random Trees (RRT)~\cite{l-rrtntpp-1998}. 
This method iteratively grows a tree, $T$, from $q_{start}$ to $q_{goal}$.
During each iteration, a random configuration $q_{rand}$ is sampled. 
We then find $q_{near}$, the configuration in $T$ closest to $q_{rand}$. 
\red{$T$ is then extended a maximum distance $\Delta$ from $q_{near}$ in the direction of $q_{rand}$.}
RRTs exhibit a Voronoi bias that results in the rapid exploration of the \cspace{} and makes RRTs an efficient way to handle single-query motion planning problems. 
RRT variants have been developed to improve performance in the presence of narrow passages~\cite{rtla-obrrt-06,yjsl-ddreecsd-05}.

\subsection{Multi-robot Motion Planning}

Multi-robot motion planning consists of finding \red{valid} paths for a set of robots between their respective starts and goals.
\red{
The composite space is the Cartesian product of each of the $n$ individual robots' \cspaces: $C_{composite} = C_1 \times C_2 \times ... \times C_n$
where $C_i$ represents the \cspace{} of robot $i$.}
A composite configuration consists of values for each robot's \dofs{}.
The composite free space is made up of all \red{valid configurations such that no robot is in collision with another robot.}
The MRMP problem can be formulated as finding a continuous path through the composite free space.

Table \ref{tab:overview} \red{compares} select MRMP approaches.
Decoupled approaches such as Decoupled PRM~\cite{sl-uppccdpmrs-2002} plan individual robot paths in their own decoupled \cspaces{} \red{and thus do not offer completeness or optimality guarantees}.
The lack of \red{coordination} degrades their performance in narrow passages, where inter-robot collisions \red{may not be possible to avoid along the individual paths only using velocity tuning.}

Coupled methods such as Composite PRM~\cite{sl-uppccdpmrs-2002} and Composite RRT~\cite{l-rrtntpp-1998} plan directly within the composite \cspace.
Other composite methods (e.g., MRdRRT~\cite{ssh-faniaehdrfeoirimm-16}) build individual robot roadmaps, then search an implicit composite roadmap.
These methods maintain the probabilistic completeness of the single-robot methods they use.
\red{However, due to the decoupled individual roadmap construction, they lack the level of coordination required to efficiently find paths in congested environments with narrow passages.}

\red{Hybrid methods such as CBS-MP~\cite{smsa-rmmpucs-21}, MAPF/C~\cite{hpksga-tpqs-2018}, and M*~\cite{wc-sefmpp-15} seek to leverage the strengths of both coupled and decoupled methods.}
For example, CBS-MP~\cite{smsa-rmmpucs-21} plans individual robot paths in their decoupled \cspaces{} and then reconciles the paths in the composite \cspace.
In the worst case, these methods will explore the whole composite \cspace, but on average, the runtimes are comparable to decoupled methods while providing varying levels of probabilistic completeness and representation optimality guarantees. These methods are generally not well suited for environments with narrow passages \red{due to the computational effort expended transitioning the planner from decoupled to the high level of coordination required.}
\red{Here, we propose a method for composite-space RRT construction while leveraging topological guidance to improve performance in environments with narrow passages.}

\begin{table}
\centering
\caption{Overview of Related Methods}
\label{tab:overview}
\scalebox{0.8}{
\begin{tabular}{|l|l|c|c|c|} 
\hline
\multicolumn{1}{|c|}{\textbf{Algorithm}} & \multicolumn{1}{c|}{\textbf{Coordination}} & \textbf{Scalable} & \begin{tabular}[c]{@{}c@{}}\textbf{Probabilistic}\\\textbf{Completeness}\end{tabular} & \begin{tabular}[c]{@{}c@{}}\textbf{Narrow}\\\textbf{Passages}\end{tabular}  \\ 
\hline
Decoupled PRM~\cite{sl-uppccdpmrs-2002} & Decoupled & X & & \\ \hline
Composite RRT~\cite{l-rrtntpp-1998} & Coupled & & X & \\ \hline
Composite PRM~\cite{sl-uppccdpmrs-2002} & Coupled & & X & \\ \hline
MRdRRT~\cite{ssh-faniaehdrfeoirimm-16} & Coupled & X & X & \\ \hline
CBS-MP~\cite{smsa-rmmpucs-21} & Hybrid & X & X & \\ \hline
MAPF/C~\cite{hpksga-tpqs-2018} & Hybrid & X & X & \\ \hline
\textbf{CDR-RRT [Ours]} & Coupled & X & X & X \\ \hline
\end{tabular}
}
\end{table}

\subsection{Guided Motion Planning}
\red{
Topological guidance has not been well explored in the composite space; however, some hybrid methods have explored using topological information to construct decoupled roadmaps.
Ryan \cite{r-cbmrpp-2010} decomposes the workspace into halls, represented by singly-linked chains of vertices, and open spaces, represented by fully-connected subgraphs.
Yu et al. \cite{Yu2018} propose a method for roadmap construction via overlaying a lattice structure onto the workspace. They show that this method leads to efficient path planning for large groups of robots in relatively open environments.
}

Several single-robot motion planning strategies have been proposed to \red{adapt planning to} the workspace.
The Feature Sensitive Motion Planning Strategy \cite{mtpra-mlafsmp-04} \red{attempts to subdivide} the environment into homogeneous workspace regions that are planned in individually, 
adapting roadmap construction to local features.
The individual roadmaps are merged into a complete roadmap of the planning space. This strategy allows the planner to use resources efficiently.

Workspace Decomposition Strategies \cite{kh-wisprp-04, bo-uwignsprp-04, kh-wco-06} help \red{concentrate} planning \red{in} narrow areas of the workspace. SyClop \cite{pkv-mpdsclp-10} uses an RRT to sample frontier decomposition regions.
A User-Guided Planning  Strategy \cite{dsja-arbsfcrc-14} allows the user to define and manipulate workspace sampling regions that the planner explores in real-time. The planner relies on the user’s intuition to \red{identify} narrow passages and find paths faster.

Skeleton-based strategies leverage the topology of the workspace \red{using an embedded graph (Fig. \ref{fig:init-skel}) that is homotopy equivalent to the workspace.}
All points in the workspace can be smoothly collapsed to the skeleton~\cite{blk-tcsrp-2012}.
Skeleton edges describe contiguous volumes of the free workspace (e.g., tunnels or rooms) and vertices represent connections between these volumes. 
\red{Given the environment, the skeleton is precomputed and may be used for multiple queries and different types of robots.
Examples include medial axis skeletons~\cite{Blum_1967_6755} in 2D and mean curvature skeletons~\cite{t-mcs-12} in 3D. Skeletons are generally quick to compute. The medial axis skeleton, for example, can be computed in $O(n \log n)$ time where $n$ is the number of obstacle edges~\cite{l-matoaps-1982}.} 

\drprm{} (DR-PRM)~\cite{suda-tgrcdrs-20}, initiates local components at the vertices of a skeleton, expands them along adjacent edges, then merges them to form a complete roadmap. 
\red{Hierarchical Annotated Skeleton Planning \cite{uyma-hpwakg-2022} extends DR-PRM by relaxing its reliance on skeleton edges over time.}
We describe the single-query counterpart of DR-PRM, \drbrrt{} (DR-RRT), in detail in Section \ref{sec:dr-rrt} since we extend our method from it. 
These methods show the advantage of using workspace information to guide planning in \cspace{} when they are closely related; however, they are constrained to single-robot settings.

\subsection{Dynamic Region-biased RRT}
\label{sec:dr-rrt}
DR-RRT~\cite{dsba-drbrrt-16} \red{(Alg.~\ref{alg:dynamicregions})} grows an RRT while constraining sampling within regions that advance along a skeleton.

\subsubsection{Query Skeleton}

\begin{figure}
    \centering
    \hfill
    \subfigure[Initial Skeleton]{
        \includegraphics[width=0.18\textwidth]{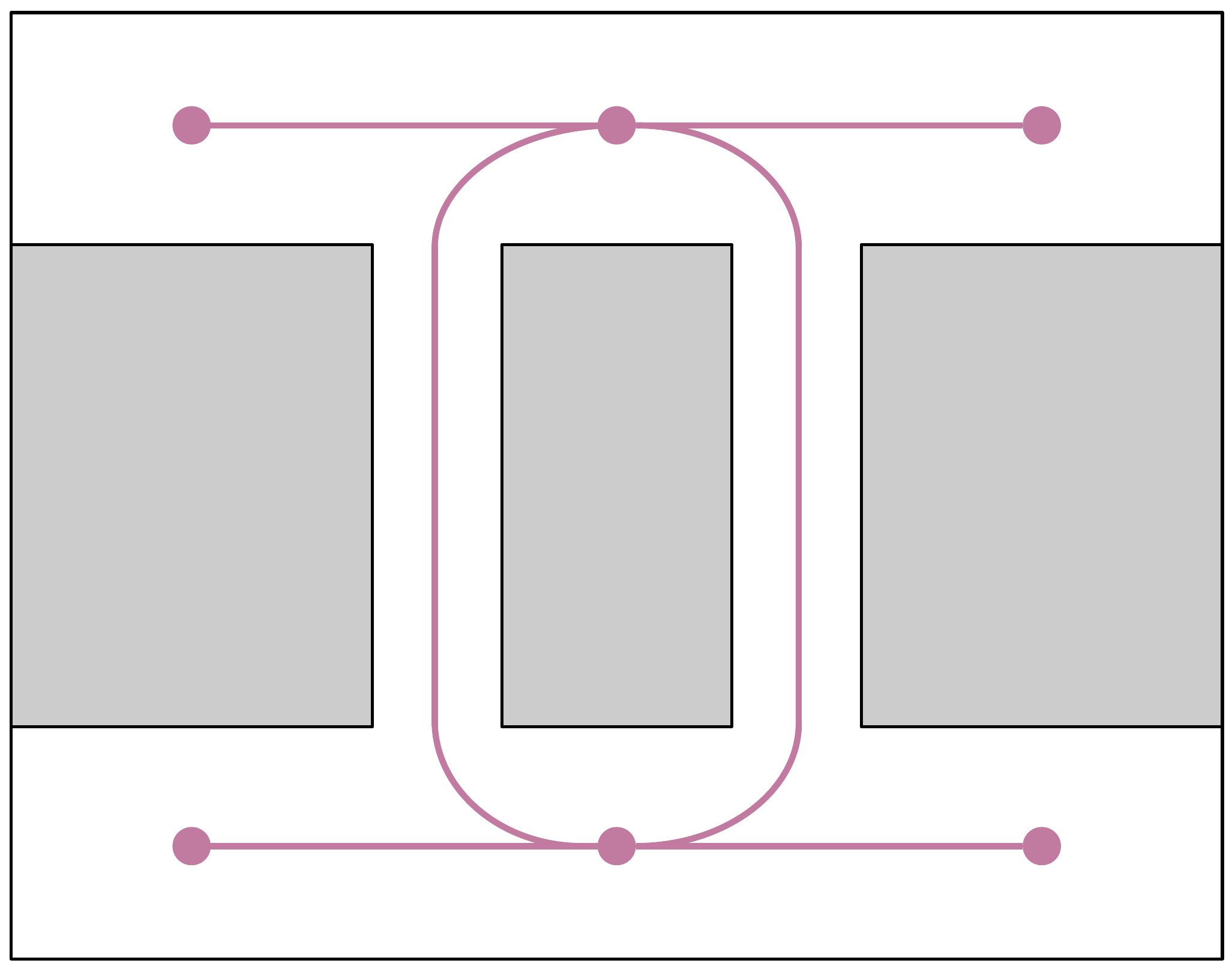}
        \label{fig:init-skel}
    }
    \hfill
    \subfigure[Query Skeleton]{
        \includegraphics[width=0.18\textwidth]{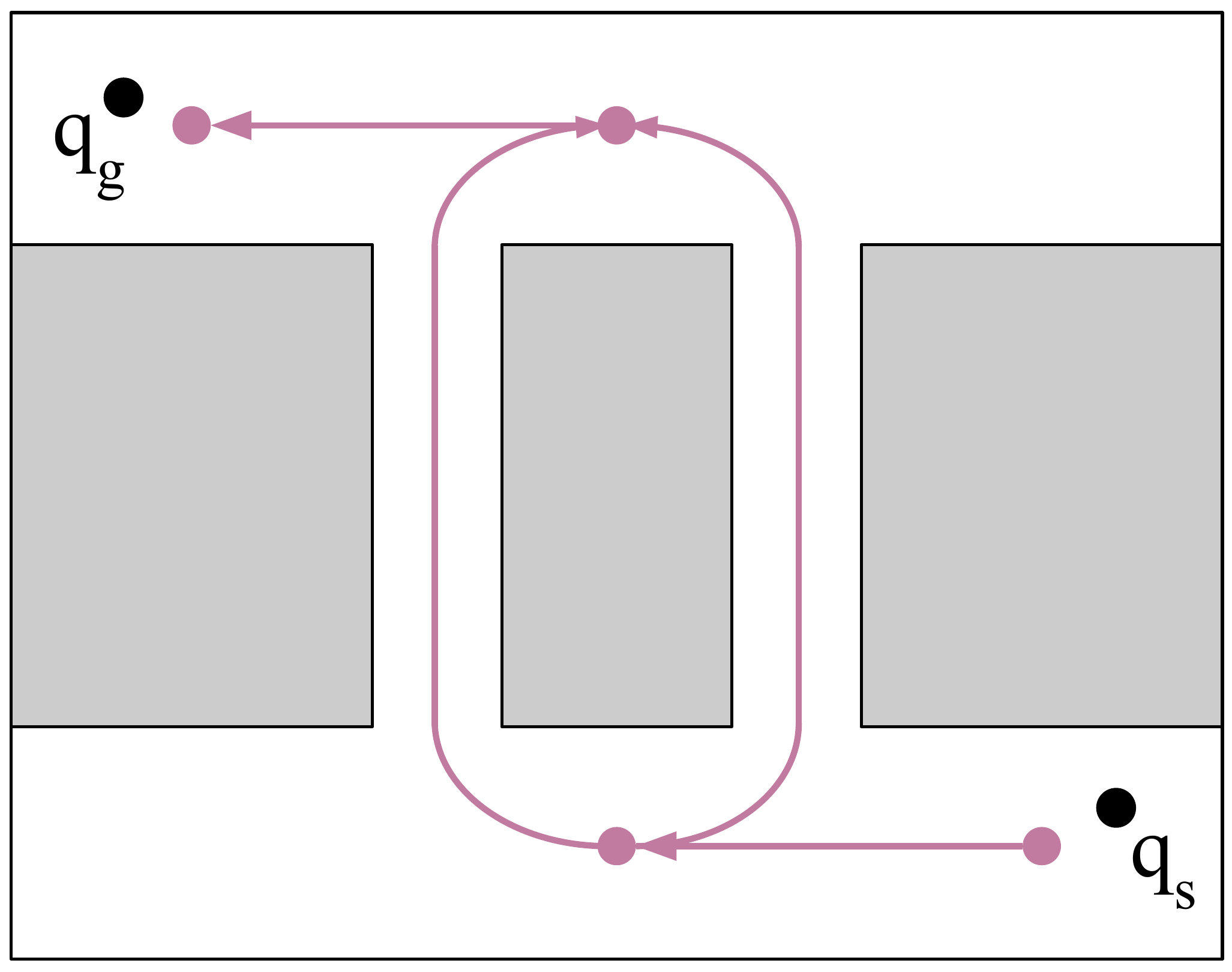}
        \label{fig:query-skel}
    }
    \hfill
    \hfill
    \caption{An example workspace skeleton (a) and query skeleton (b). The query skeleton is a directed and pruned skeleton that only contains edges along a path from $q_{start}$ to $q_{goal}$.}
    \label{fig:dr-rrt-skeletons}
\end{figure}

Algorithm \ref{alg:dynamicregions} creates two skeletons: one for the entire workspace (line \ref{alg:init-skel}), and 
the \textit{query skeleton} (Fig. \ref{fig:query-skel}; line \ref{alg:query-skel}), which
retains only edges that are along a path from the start to the goal in the workspace. 

\begin{algorithm}[t]
\small
  \caption{\drbrrt}
  \label{alg:dynamicregions}
  \begin{algorithmic}[1]
    \INPUT Environment $e$ and a Query $\{q_s, q_g\}$
    \OUTPUT Tree $T$
    \State $W \gets \CALL{ComputeWorkspaceSkeleton}(e)$ \label{alg:init-skel}
    \State $S \gets \CALL{ComputeQuerySkeleton}(W, q_s, q_g)$ \label{alg:query-skel}
    \State $T \gets (\emptyset, \emptyset)$
    \State $R \gets \CALL{InitialRegions}(S, q_s)$ \label{alg:init-regions}
    \While {$\neg (q_g \in T)$}
    \State $\CALL{RegionBiasedRRTGrowth}(T, S, R)$
    \EndWhile
    \State \Return $T$
  \end{algorithmic}
\end{algorithm}

\begin{algorithm}[h]
\small
  \caption{Region-biased \rrt\ Growth}
  \label{alg:rbrrtg}
  \begin{algorithmic}[1]
    \INPUT Tree $T$, \qskeleton\ $S$, Regions $R$, region radius $\eta$, maximum for failed extension attempts $\tau$
    \State $r \gets \CALL{SelectRegion}(R)$ \label{alg:select-region}
    \State $q_{rand} \gets r.\CALL{GetRandomCfg}()$
    \State $q_{near} \gets \CALL{NearestNeighbor}(T, q_{rand})$ \label{alg:qrand}
    \State $q_{new} \gets \CALL{Extend}(q_{near}, q_{rand}, \Delta)$ \label{alg:qnew}
    \If{$r = \emptyset$} \Comment{The entire environment was chosen}
    \State \Return
    \EndIf
    \If {$q_{new}.\CALL{ExtensionSucceeded}()$} \label{alg:success-rate}
    \State $r.\CALL{IncrementSuccesses()}$
    \Else
    \State $r.\CALL{IncrementFailures()}$ \label{alg:end-success-rate}
    \EndIf
    \For {$r \in R$} \label{alg:for-advance}
    \While {$r.\CALL{InRegion}(q_{new})$}
    \State $r.\CALL{AdvanceAlongSkeletonEdge}()$ \label{alg:end-for-advance}
    \If {$r.\CALL{AtEndOfSkeletonEdge}()$} \label{alg:delete-end}
    \State $R \gets R \setminus \{r\}$ \label{alg:end-delete-end}
    \EndIf
    \EndWhile
    \EndFor
    \If {$r.\CALL{NumFailures}() > \tau$} \label{alg:fail-start}
    \State $R \gets R \setminus \{r\}$ \label{alg:fail-end}
    \EndIf
    \ForAll {$v \in S.\CALL{UnexploredVertices}()$} \label{alg:new-regions}
    \If {$\delta(v, q_{new}) < \eta$}
    \State $R \gets R \cup \CALL{NewRegion(v)}$
    \State $S.\CALL{MarkExplored}(v)$ \label{alg:end-new-regions}
    \EndIf
    \EndFor
  \end{algorithmic}
\end{algorithm}

\subsubsection{Sampling Regions}
Sampling regions will advance along the edges of the query skeleton, 
\red{using the skeleton, a solution in the workspace, to guide construction of an RRT, a solution in the \cspace{}.}
A \textit{region} is a bounded volume in the workspace, e.g., a bounding sphere. 
Construction of the RRT begins by initializing the first region centered on the skeleton vertex closest to $q_{start}$ (line \ref{alg:init-regions}).
During each iteration, \red{Algorithm \ref{alg:rbrrtg} selects} a region to guide sampling (line \ref{alg:select-region}). 
The probability of selecting a region is proportional to its \textit{extension success rate}. \red{To maintain probabilistic completeness,} with a small probability, the entire environment is chosen \red{(see Sec.~\ref{sec:theory})}.
A random configuration $q_{rand}$ is selected from this region to grow the tree toward (line \ref{alg:qrand}). The algorithm then proceeds as a general RRT by attempting to extend the tree to $q_{new}$ (line \ref{alg:qnew}). The extension success rate is updated based on the outcome of this attempt (lines \ref{alg:success-rate}-\ref{alg:end-success-rate}).

\subsubsection{Region Advancement}
Once a configuration has been added to the tree, all regions that are in contact with $q_{new}$ are advanced forward along their skeleton edges until they leave $q_{new}$ behind 
(Alg.~\ref{alg:rbrrtg}, lines \ref{alg:for-advance}-\ref{alg:end-for-advance}). 
Any region that reaches the end of its edge \red{or exceeds the maximum number of extension failures} is deleted (lines \ref{alg:delete-end}-\ref{alg:fail-end}). Then, new regions are created on each unexplored skeleton vertex that is within a small distance of $q_{new}$ (lines \ref{alg:new-regions}-\ref{alg:end-new-regions}). This cycle of region selection, tree extension, region advancement, and region creation continues until the tree extends to $q_{goal}$.



\section{COMPOSITE DYNAMIC REGION-BIASED RRT}

\red{In this paper,} we extend DR-RRT to multi-robot systems to propose a new method, Composite Dynamic Region-biased RRT (CDR-RRT).
\red{We limit the exploration of the composite space to areas that are likely to yield a solution because} of the exponential size of the search space.
\red{
We do this by developing composite analogs for workspace skeletons and regions that allow for coupled multi-robot motion planning while leveraging topological guidance as in DR-RRT.
We leverage lazy construction of the composite skeleton as we exploit a greedy heuristic to search the composite space.
}

\subsection{Composite Skeleton}
\label{sec:comp-skel}
\begin{figure}[t]
    \centering
    \hfill
    \subfigure[Robot 1]{
        \includegraphics[width=0.14\textwidth]{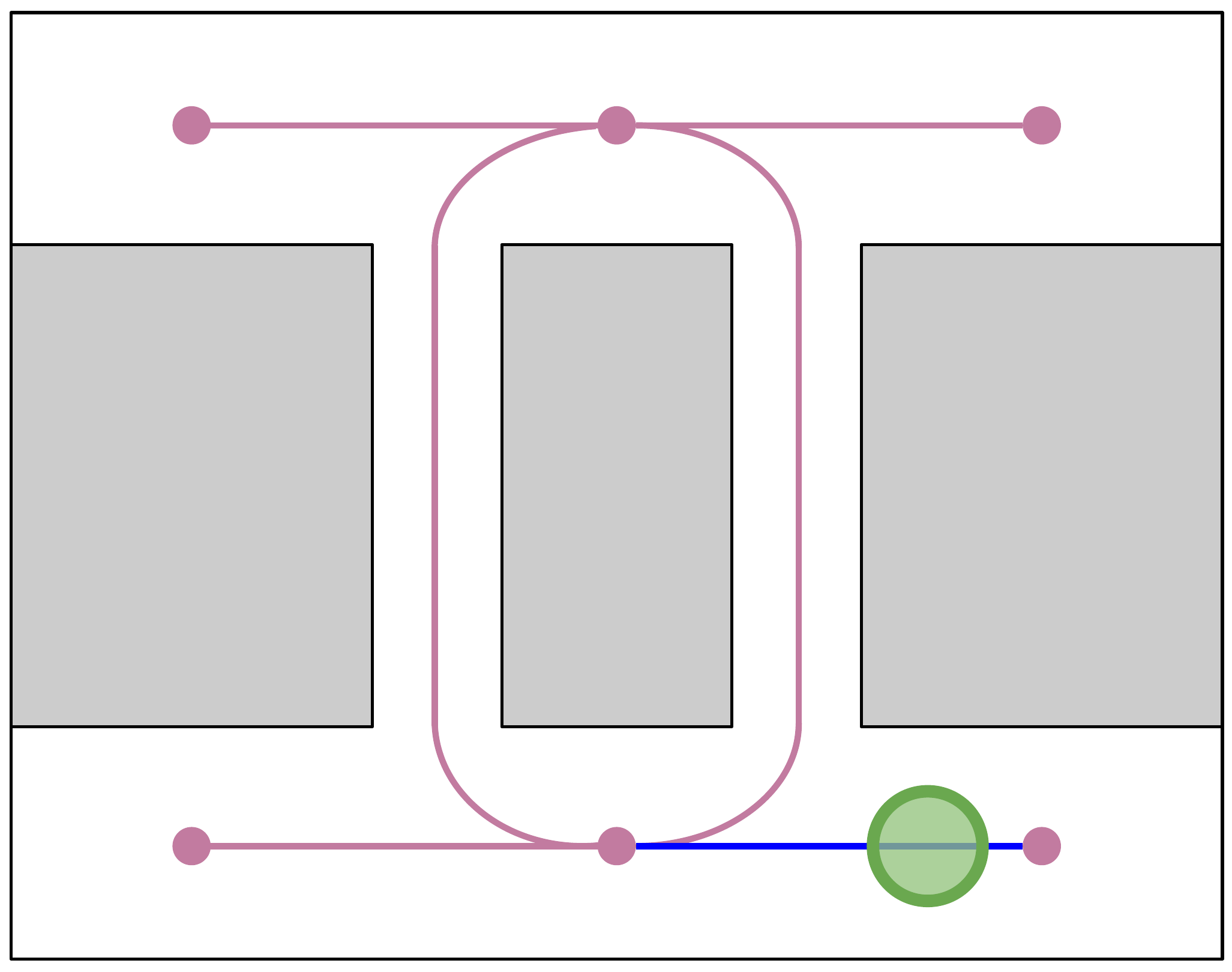}
        \label{fig:comp-edge-1}
    }
    \hfill
    \subfigure[Robot 2]{
        \includegraphics[width=0.14\textwidth]{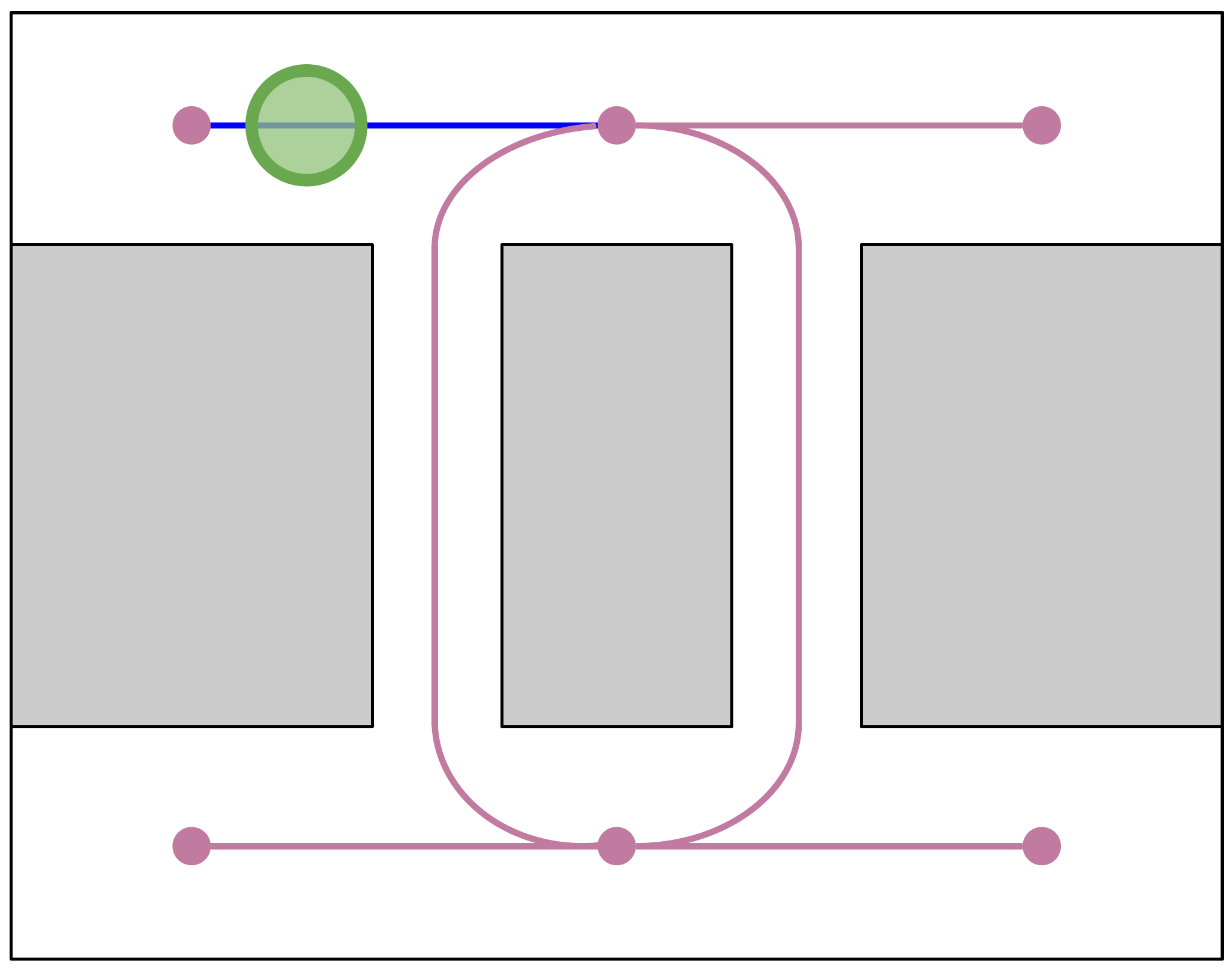}
        \label{fig:comp-edge-2}
    }
    \hfill
    \subfigure[Robot 3]{
        \includegraphics[width=0.14\textwidth]{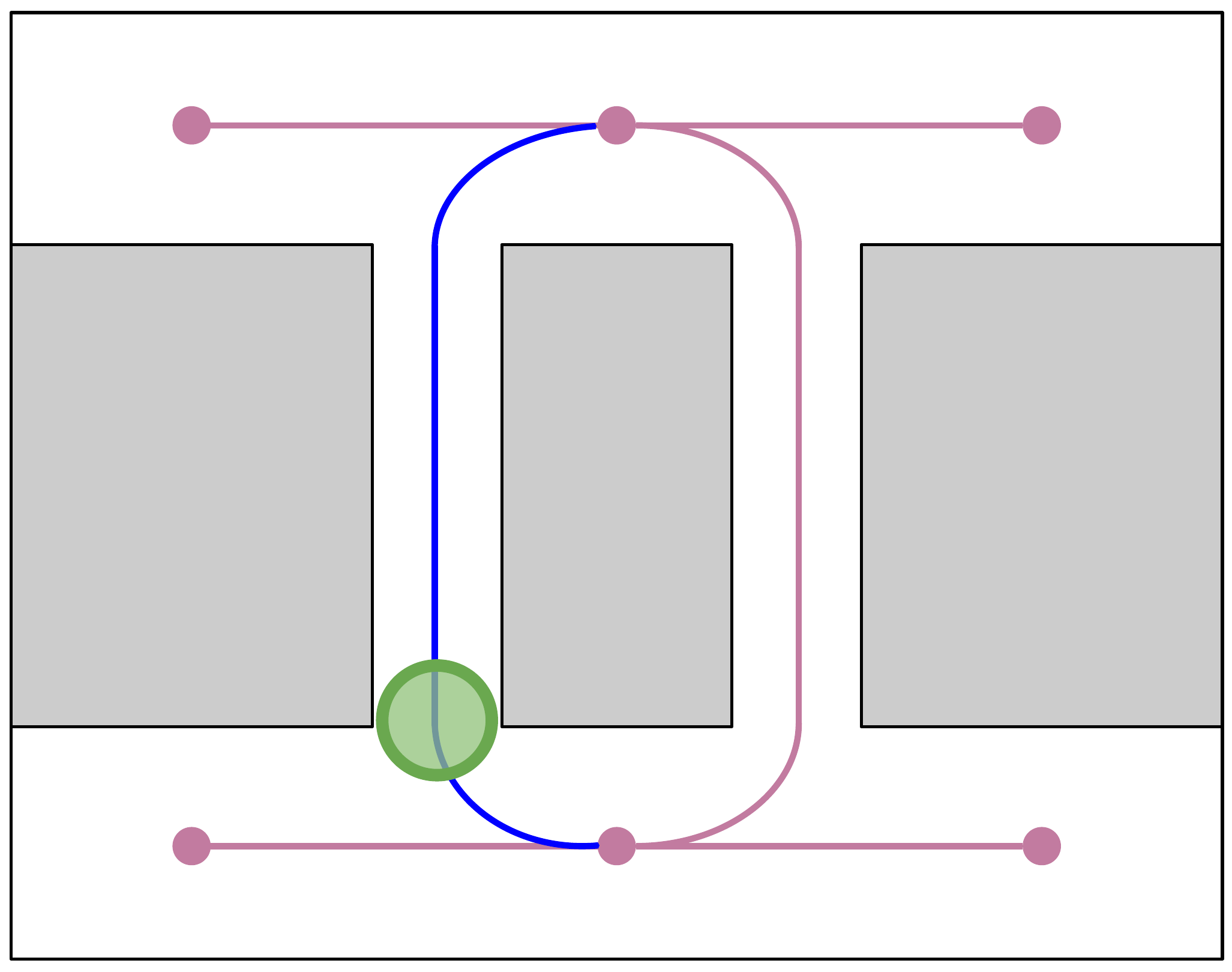}
        \label{fig:comp-edge-3}
    }
    \hfill
    \hfill
    
    \caption{
    An example of a composite edge and a composite region. 
    Each of the three robots has its own workspace skeleton. 
    Obstacles are gray, and the workspace skeletons are purple. 
    Three skeleton edges that comprise a composite edge are shown in blue.
    All possible combinations of such edges make up the composite skeleton. 
    Three individual regions that make up a composite region along this composite edge are shown in green.
    }
    \label{fig:composite-edge}
\end{figure}

\begin{algorithm}[t]
  \small
  \caption{\red{Composite Region-biased \rrt\ Growth}}
  \label{alg:crbrrtg}
  \begin{algorithmic}[1]
    \INPUT Tree $T$, Composite Skeleton $S$, Region $r$, Failed Vertex and Edge Constraints $C$, Maximum for Failed Extension Attempts $\tau$
    \State $Bound \gets \CALL{SelectRegionOrWholeEnvironment}()$\label{alg:bounds}
    \State $q_{rand} \gets r.\CALL{GetRandomCompositeCfg}(Bound)$
    \State $q_{near} \gets \CALL{NearestNeighbor}(T, q_{rand})$
    \State $q_{new} \gets \CALL{Extend}(q_{near}, q_{rand}, \Delta)$
    \If{$Bound = \CALL{WholeEnvironment}$}
    \State \Return
    \EndIf
    \If {$q_{new}.\CALL{ExtensionSucceeded}()$}
    \State $r.\CALL{IncrementSuccesses()}$
    \Else
    \State $r.\CALL{IncrementFailures()}$
    \EndIf
    \While {$r.\CALL{InCompositeRegion}(q_{new})$}
    \State $r.\CALL{AdvanceAlongCompositeSkeletonEdge}()$\label{alg:end-adv-siblings}
    \If {$r.\CALL{AtEndOfCompositeSkeletonEdge}()$}\label{alg:delete-comp-region}
    \State $V_t \gets r.\CALL{CompositeTargetVertex()}$
    \State $r \gets S.\CALL{GrowCompositeSkeleton}(V_t, C)$ \label{alg:delete-grow-skel}
    \EndIf
    \EndWhile
    \If {$r.\CALL{NumFailures}() > \tau$} \label{alg:priority-queue}
    \State $C \gets C \cup \{r.\CALL{Edge\}}$
    \State $V_s \gets r.\CALL{CompositeSourceVertex}()$
    \If {$V_s.\CALL{ExceededMaximumFailures}()$}
    \State $C \gets C \cup \{V_s\}$
    \State $V_s \gets V_s.\CALL{GetPredecessor}()$
    \EndIf
    \State $r \gets S.\CALL{GrowCompositeSkeleton}(V_s, C)$ \label{alg:replace-region}
    \EndIf
  \end{algorithmic}
\end{algorithm}

\red{As the composite space is the Cartesian product of each robot's \cspace{},
the composite skeleton (Fig.~\ref{fig:composite-edge}) is the Cartesian product of the workspace skeleton for each of the $n$ robots.
It consists of composite vertices and edges which respectively represent a set of $n$ vertices or edges in the workspace skeleton where each of the $n$ robots lies. We avoid the exponential expansion associated with the computation of the full composite skeleton graph by using local, on-demand construction.}
A composite region is made up of $n$ individual sampling regions, one in each robot's workspace.

Computing a composite query skeleton, which is a directed and pruned version of the composite skeleton, requires an explicit computation of the composite skeleton. 
\red{
Instead, we heuristically construct and search the composite skeleton one edge at a time and only consider edges likely to be along a feasible low-cost path from $q_{start}$ to $q_{goal}$. 
We discuss our heuristic to capture these edges in Section \ref{section:cbs-heur}.
}

\subsection{Guided Composite RRT Construction}

\begin{algorithm}[t]
  \small
  \caption{\red{Grow Composite Skeleton}}
  \label{alg:growskeleton}
  \begin{algorithmic}[1]
    \INPUT Composite Skeleton $S$, Composite Source Vertex $V$, Failed Vertex and Edge Constraints $C$
    \While{$V.\CALL{ExceededGrowthAttempts}()$}
    \State $V \gets V.\CALL{GetPredecessor}()$ \label{alg:predecessor}
    \EndWhile
    \State $Paths \gets \CALL{MAPFSolution}(V, C)$ \label{alg:mapf}
    \State $E \gets \CALL{ExtractFirstCompositeEdge}(Paths)$ \label{alg:extract}
    \State $S.\CALL{AddEdge}(E)$
    \State \Return $\CALL{NewRegion}(E)$ \label{alg:newregion}
  \end{algorithmic}
\end{algorithm}

\red{
To begin RRT construction, we compute the first composite skeleton edge to explore.
Section \ref{section:cbs-heur} discusses the construction of composite skeleton edges by growing the composite skeleton from a source composite vertex.}
CDR-RRT then proceeds as DR-RRT by iteratively performing \red{region-biased sampling,} RRT growth, region advancement and deletion, and new region creation until $q_{goal}$ is reached.

\begin{figure}[t]
    \centering
    \includegraphics[scale=0.50]{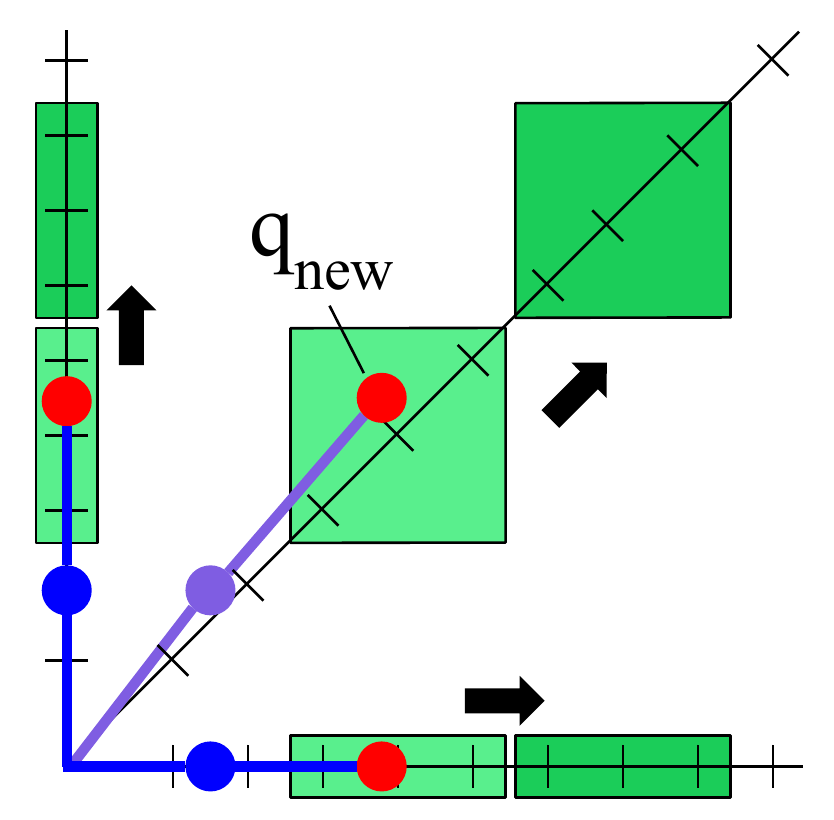}
    \caption{An example of composite region advancement for two point robots in a 1D environment. The individual robots' skeletons are shown on the axes and the composite skeleton between them. Each individual roadmap is shown in blue and the composite roadmap is in purple. Edge intermediates are shown as tick marks along the skeletons. Both individual regions advance at the same rate with respect to their individual intermediates until the composite region leaves $q_{new}$ behind at the position shown in dark green.}
    \label{fig:adv}
\end{figure}

\red{
During each iteration of CDR-RRT, Algorithm \ref{alg:crbrrtg} selects a composite region $r$.
After sampling from $r$ and extending the tree, we advance $r$ forward until it leaves $q_{new}$ behind (line \ref{alg:end-adv-siblings}).}
In DR-RRT~\cite{dsba-drbrrt-16}, as individual regions advance along skeleton edges, they are centered on intermediate points along the edges. 
Correspondingly, we create composite intermediates along composite skeleton edges.
In composite region advancement, as shown in Fig.~\ref{fig:adv}, all individual regions are advanced forward the minimum amount of intermediates such that the composite region is no longer touching $q_{new}$.

\red{
When a composite region reaches the target vertex at the end of its edge, we add an outgoing edge to the composite skeleton from this vertex and spawn a new region (Alg.~\ref{alg:growskeleton}). 
When a region surpasses $\tau$ failed extension attempts, it is deleted and replaced with a new region (line~\ref{alg:replace-region}).
}

\subsection{Multi-agent Pathfinding Heuristic}
\label{section:cbs-heur}
\red{
We use a multi-agent pathfinding (MAPF) heuristic to identify the next edge to explore given a source composite vertex, $V$ (Alg.~\ref{alg:growskeleton}).
MAPF is the discrete state space equivalent of the MRMP problem.
We use MAPF to generate a path for each robot through the workspace skeleton from $V$ to the vertex closest to each robot's goal (line \ref{alg:mapf}).}
We ensure that these individual paths are feasible by accounting for potential collisions between robots. We define the capacity of an individual skeleton edge as the minimum width between obstacles along the edge. 
If the total width of the robots traversing that edge exceeds the capacity, a conflict has occurred. These conflicts are resolved by \red{the MAPF algorithm}.

\begin{figure}[t]
    \centering
    \hfill
    \subfigure[Failed Edge]{
        \includegraphics[height=0.14\textwidth]{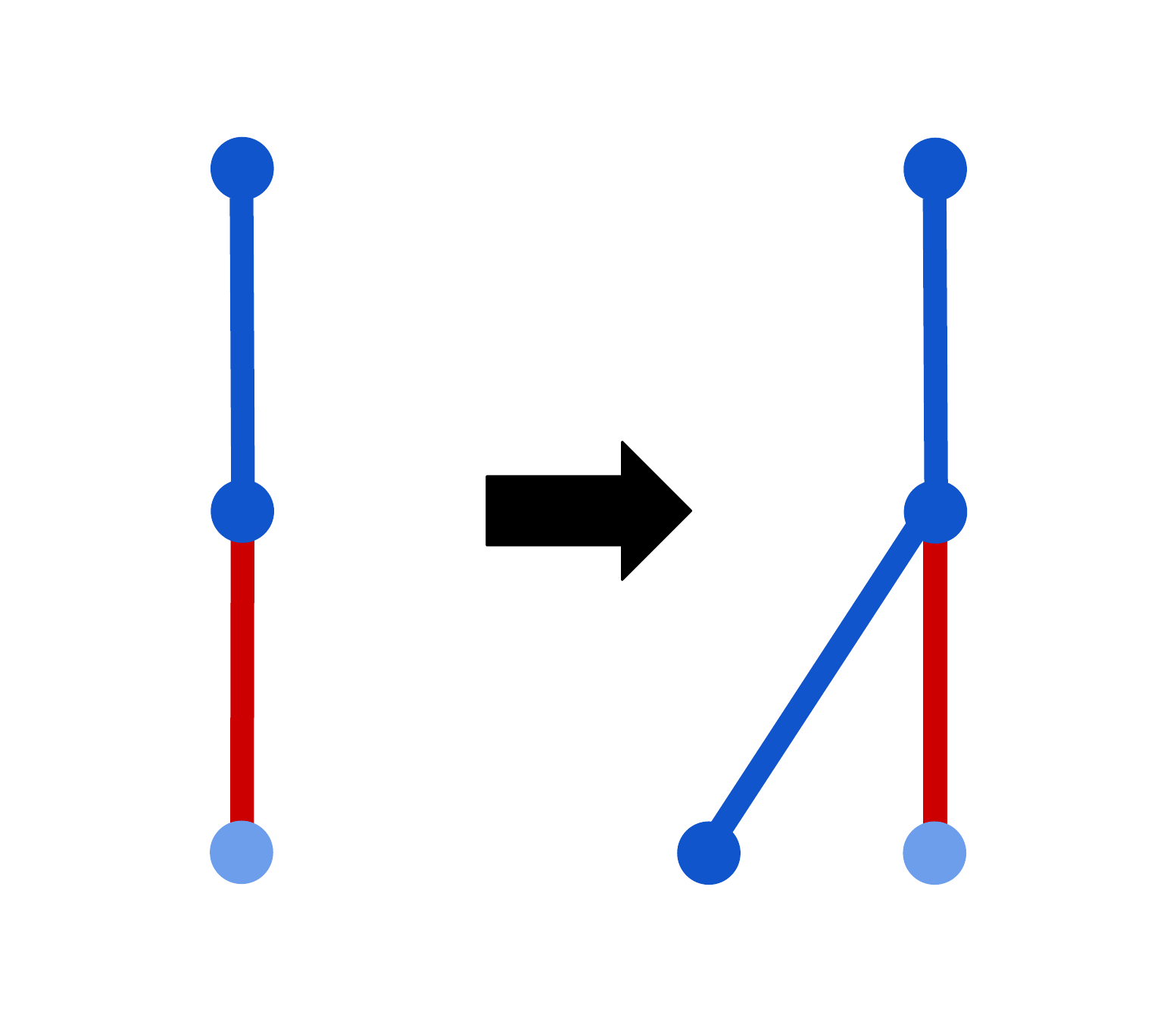}
        \label{fig:failed_comp_edge}
    }
    \hfill
    \subfigure[Failed Vertex]{
        \includegraphics[height=0.14\textwidth]{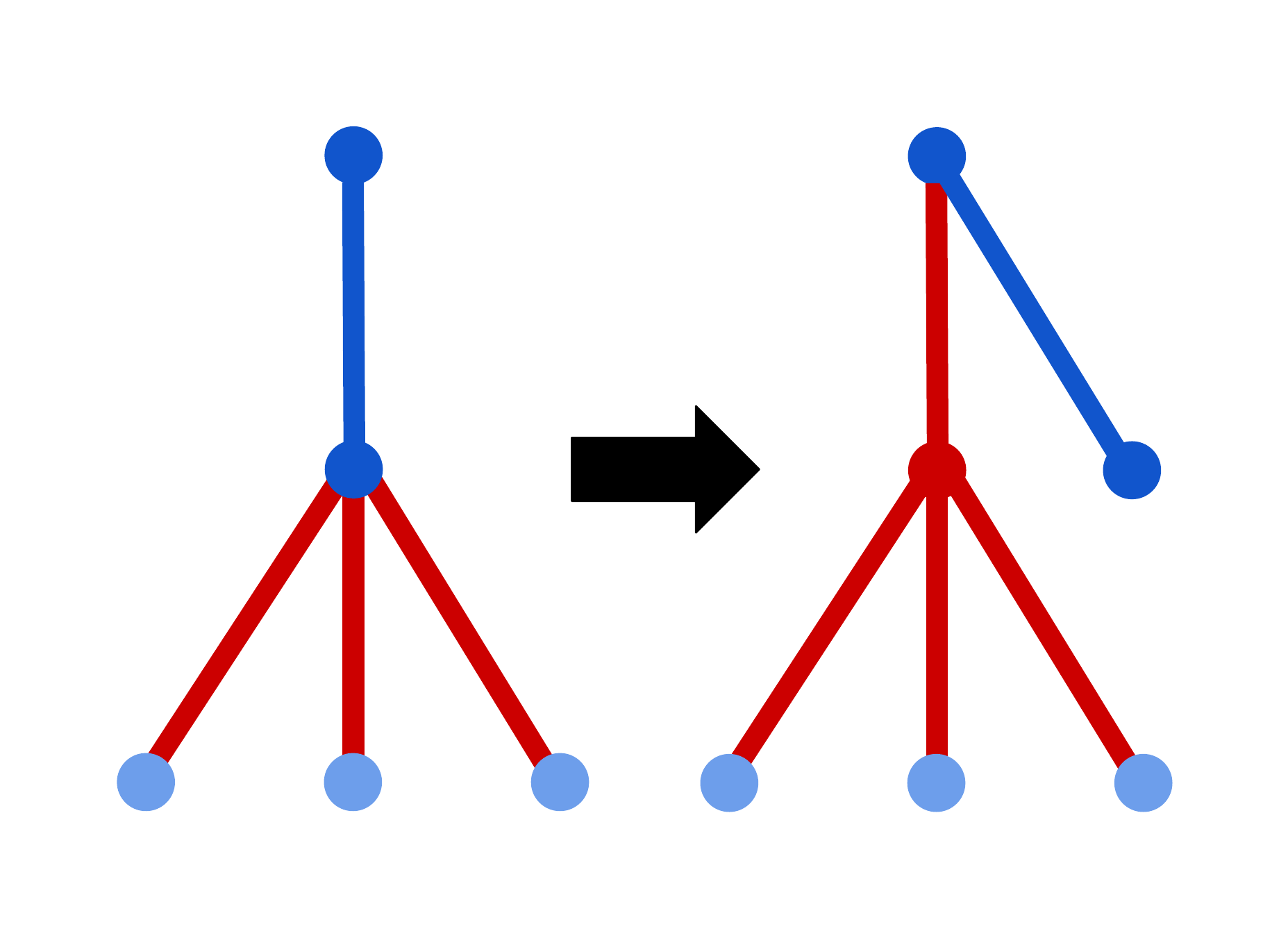}
        \label{fig:failed_comp_vertex}
    }
    \hfill
    \hfill
    
    \caption{
    \red{An example of the MAPF replanning process. (a) If a composite skeleton edge (in red) cannot be traversed, we impose a constraint that future MAPF solutions cannot contain that edge and replan the paths from the last vertex reached. (b) If the maximum number of failed outgoing edges from a vertex is exceeded, a constraint is imposed that future MAPF solutions cannot contain that vertex, and paths are replanned from its predecessor.}
    }
    \label{fig:mapf-replan}
\end{figure}

\red{
We extract composite skeleton edges from the produced MAPF solution (Alg.~\ref{alg:growskeleton}, line \ref{alg:extract}) and iteratively create a region to traverse each edge.
If a region exceeds the maximum number of extension failures traversing an edge, we consider that edge failed and impose a constraint that no further MAPF solutions can contain that composite skeleton edge (Fig.~\ref{fig:failed_comp_edge}). 
We also increment the number of failed growth attempts that each source composite skeleton vertex has seen. To avoid spending excess computation exploring a region of the composite skeleton that is unlikely to produce a path, if a vertex exceeds the maximum number of growth failures, we backtrack to its predecessor vertex (Fig.~\ref{fig:failed_comp_vertex}; line \ref{alg:predecessor}). 
}

\subsection{\red{Implementation Details}}
\red{
To generate MAPF solutions, we adapt Conflict-Based Search (CBS)~\cite{ssfs-cbsfomap-15} and Priority-Based Search (PBS)~\cite{ma2019searching}.
Both use a hierarchical approach with a low-level search to find individual paths for each agent and a high-level search to resolve conflicts between paths.
CBS finds optimal paths with respect to the makespan while PBS has been shown to achieve improved performance in scenarios where the optimal path for one robot blocks the path for other robots.}

The size of the full composite skeleton is exponential in the number of robots, so we optimize memory usage by leveraging local construction of the composite skeleton. 
We can also remove composite vertices and edges when they are no longer useful.
A composite edge is no longer useful when the composite region that traverses it has reached the end and been deleted. A composite vertex is no longer useful when all of its incoming and outgoing edges are no longer useful. 


\begin{figure}[]
    \centering
    \hfill
    \subfigure[Cross]{
        \includegraphics[width=0.14\textwidth]{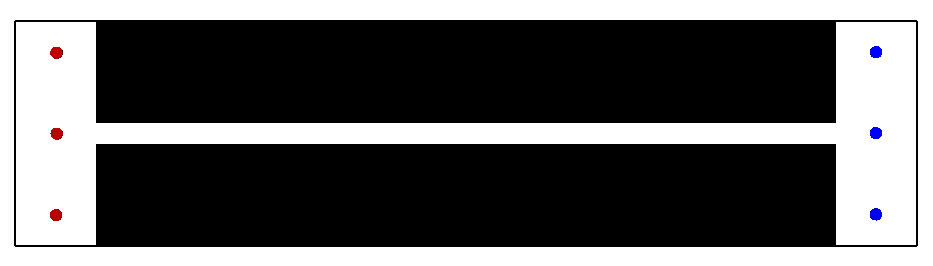}
        \label{fig:hall-cross}
    }
    \hfill
    \subfigure[Flow]{
        \includegraphics[width=0.14\textwidth]{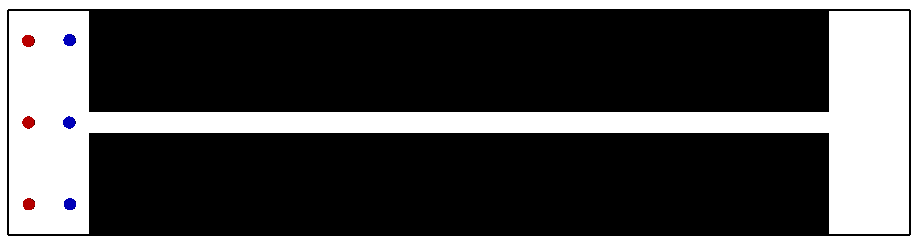}
        \label{fig:hall-flow}
    }
    \hfill
    \subfigure[Inlet]{
        \includegraphics[width=0.14\textwidth]{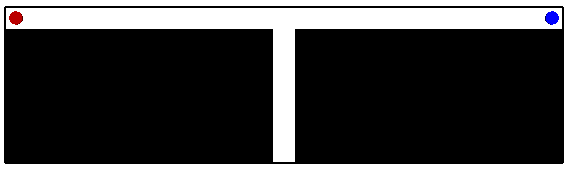}
        \label{fig:inlet}
    } \hfill \hfill \\
    \hfill
     \subfigure[Track]{
        \includegraphics[width=.14\textwidth]{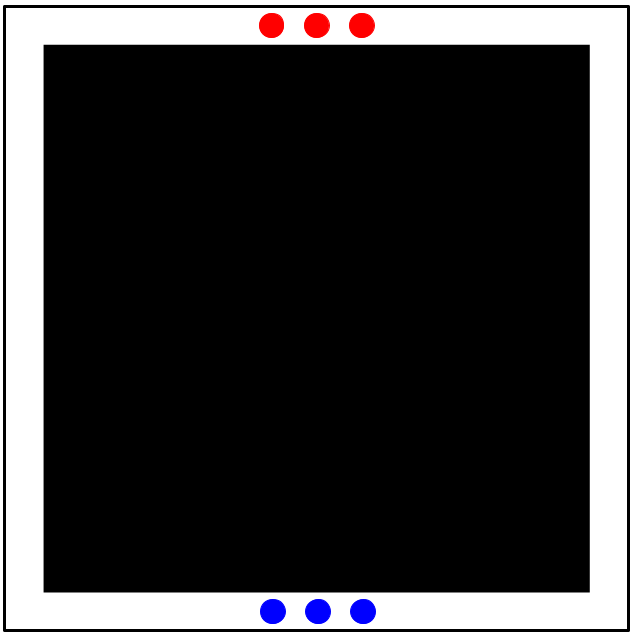}
        \label{fig:track}
    }
    \hfill
    \subfigure[Warehouse]{
       \includegraphics[width=.14\textwidth]{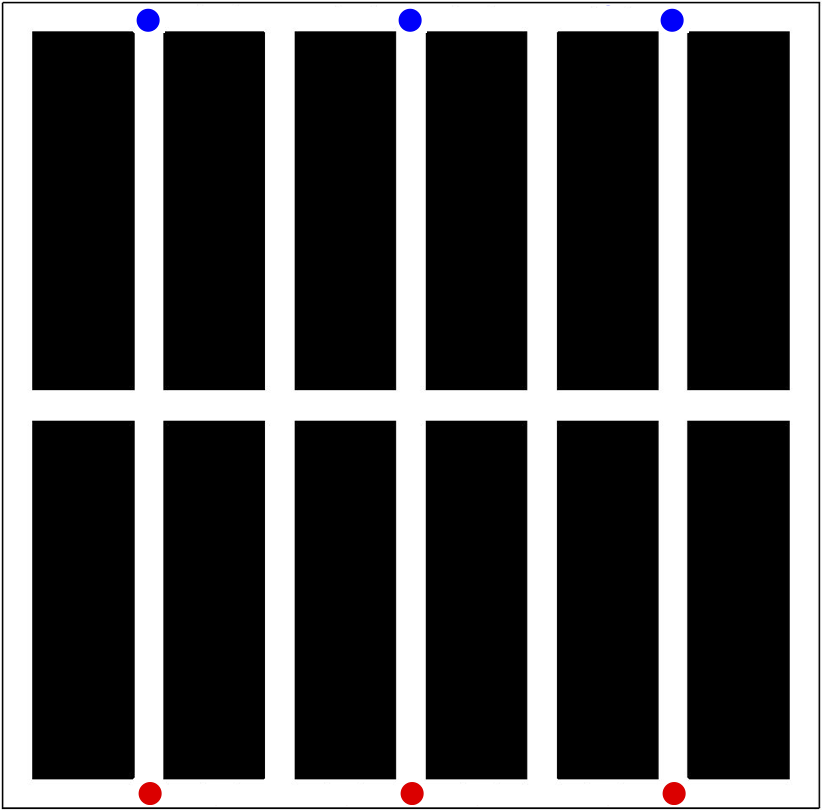}
        \label{fig:short-warehouse}
    }
    \hfill
    \hfill
\caption{\red{The Cross (a) and Flow (b) scenarios feature a single hallway through which all robots must pass. In Cross, the robots on either side must swap places. In Flow, all robots must move from one side of the hallway to the other. In the Inlet (c) scenario, the robots must swap places. One robot must move into the inlet to allow the other to pass.
The track environment (d) features a ring through which robots must move. The robots on top and bottom must swap places by moving in the same direction to avoid collision.
The Warehouse scenario (e) has three variants of aisle widths (1, 2, and 4 meters; the 1m variant is shown). The robots on top and bottom must swap places with the robot with which they are aligned vertically. }}
    \label{fig:warehouse}
\end{figure}
\begin{figure}
    \centering
    \hfill
    \subfigure[Open Cross]{
        \includegraphics[height=.14\textwidth]{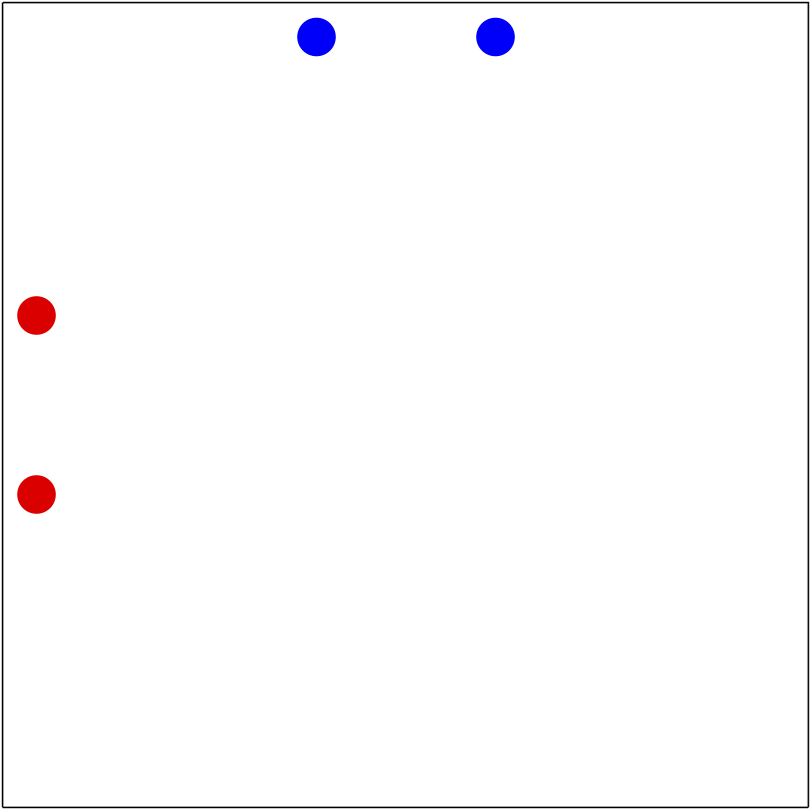}
        \label{fig:open-cross}
    }
    \hfill
    \subfigure[Maze Cross]{
        \includegraphics[height=0.14\textwidth]{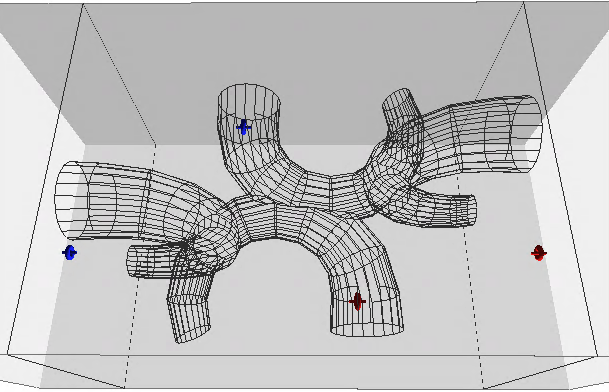}
        \label{fig:3d_maze}
    }
    \hfill
    \hfill
\caption{In the Open Cross (a) environment, the red robots must move from left to right and the blue robots from top to bottom. 
In the Maze Cross (b) environment, robots starting on the left and right must swap places by moving through the 3D tunnel structure.}
    \label{fig:cross}
\end{figure}

\begin{figure}[t!]
    \centering
    \subfigure[Hallway]{
        \includegraphics[width=.49\textwidth]{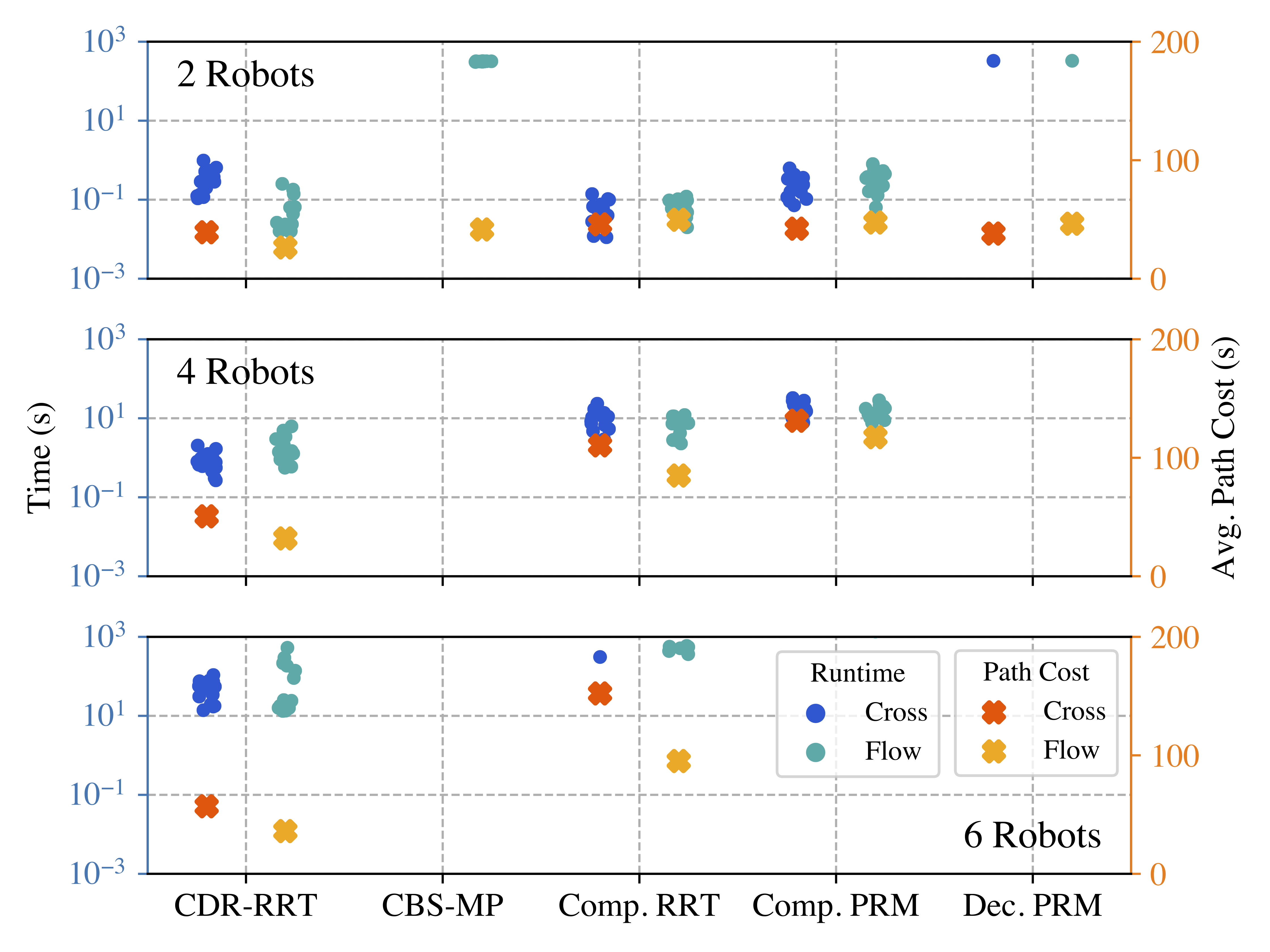}
        \label{fig:hallway-graph}
    }
    \subfigure[Inlet]{
        \includegraphics[width=.49\textwidth]{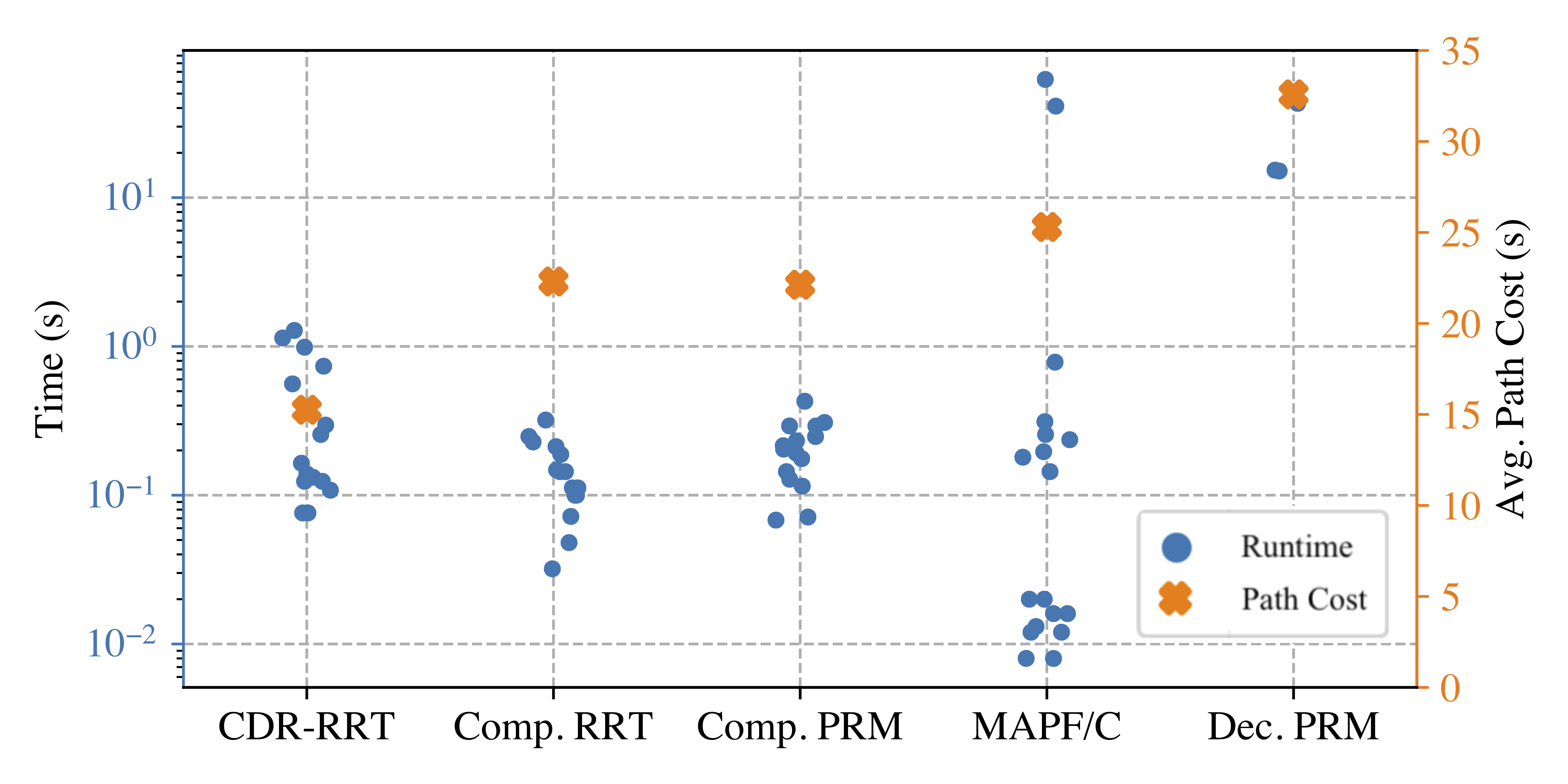}
        \label{fig:inlet-graph}
    }
    \subfigure[Track]{
        \includegraphics[width=.49\textwidth]{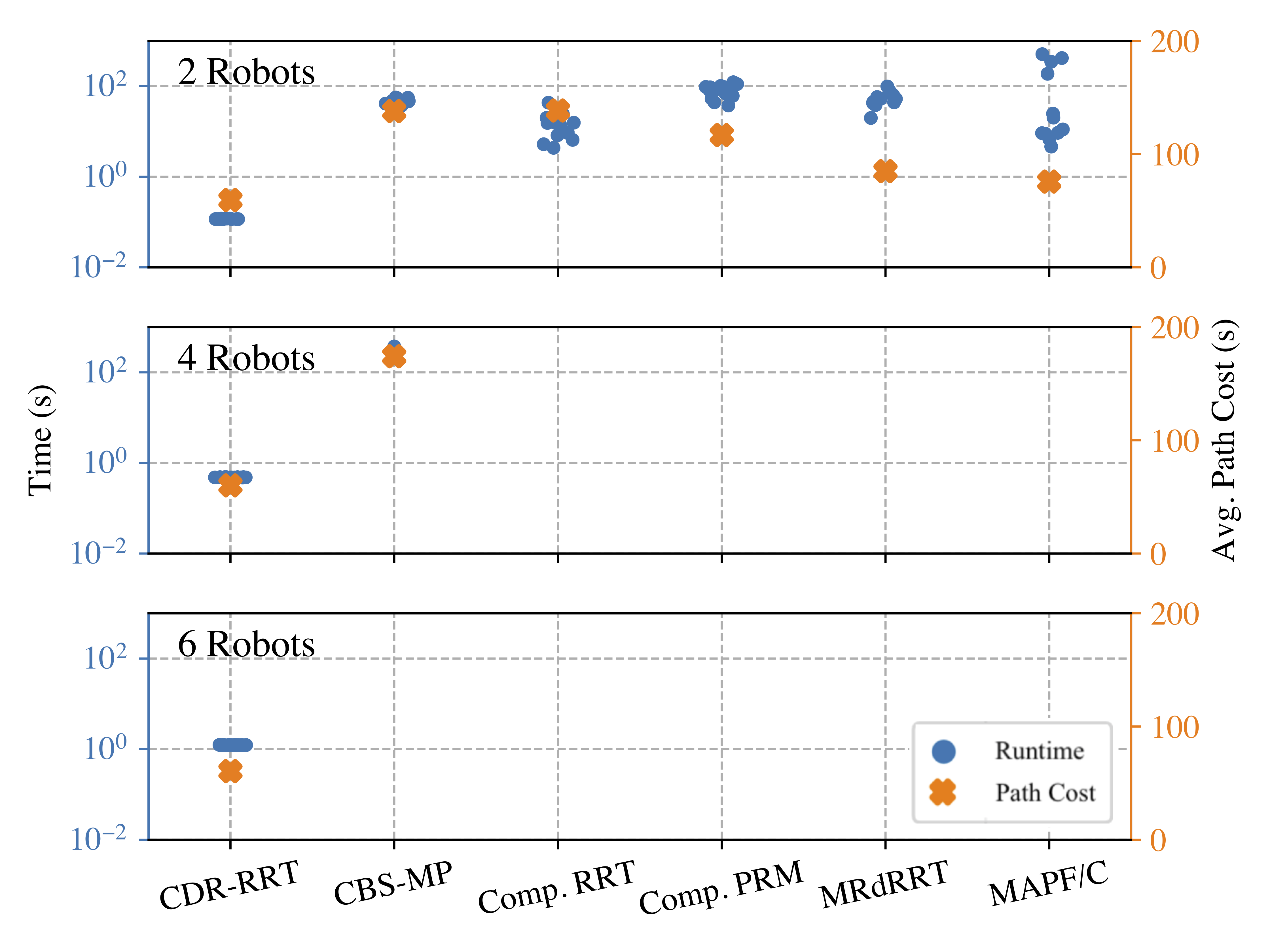}
        \label{fig:track-graph}
    }
    \caption{\red{Running time and path cost results for the Hallway (a), Inlet (b), and Track (c) environments.}}
    \label{fig:coord-graphs}
\end{figure}

\begin{figure}[t!]
    \centering
    \includegraphics[width=.49\textwidth]{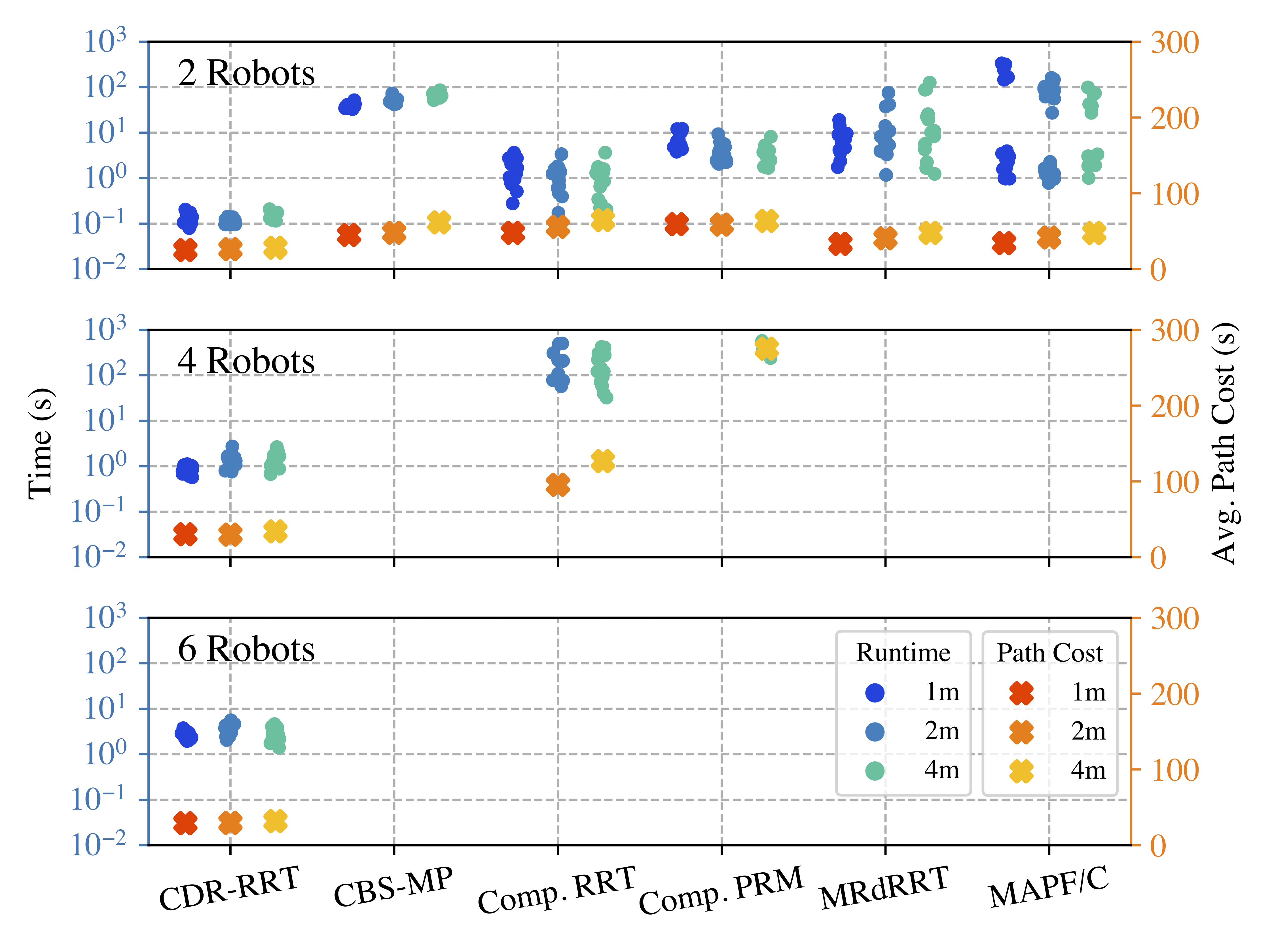}
\caption{Running time \red{and path cost} results for the Warehouse scenarios.}
    \label{fig:warehouse-both-graph}
\end{figure}

\begin{figure}[t!]
    \centering
    \includegraphics[width=0.49\textwidth]{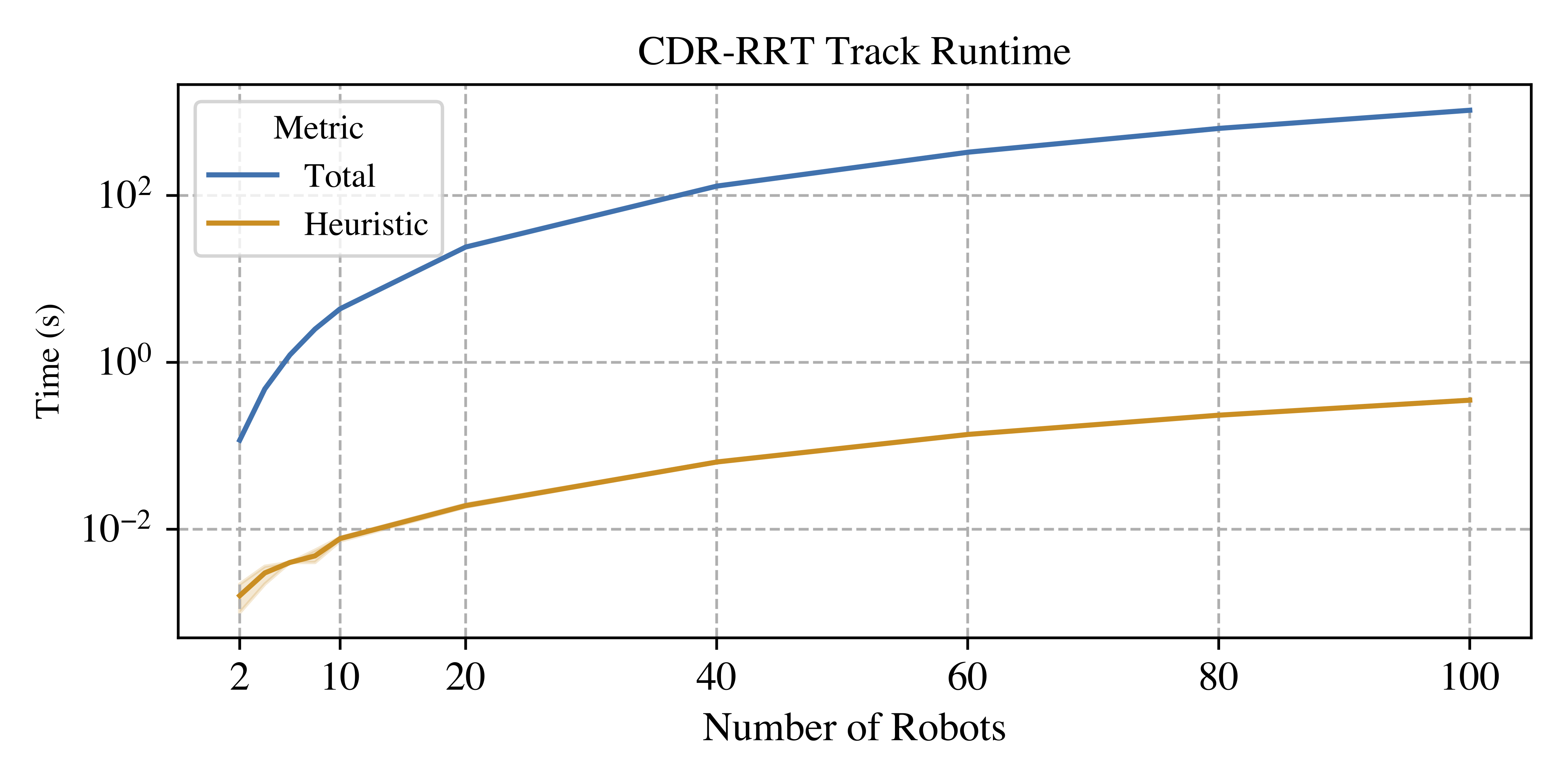}
    \caption{\red{Running time results for up to 100 robots for CDR-RRT on the Track Environment including heuristic evaluation times.}}
    \label{fig:track-cdr-rrt-graph}
\end{figure}

\begin{figure}
    \centering
    \subfigure[Open Cross]{
        \includegraphics[width=.49\textwidth]{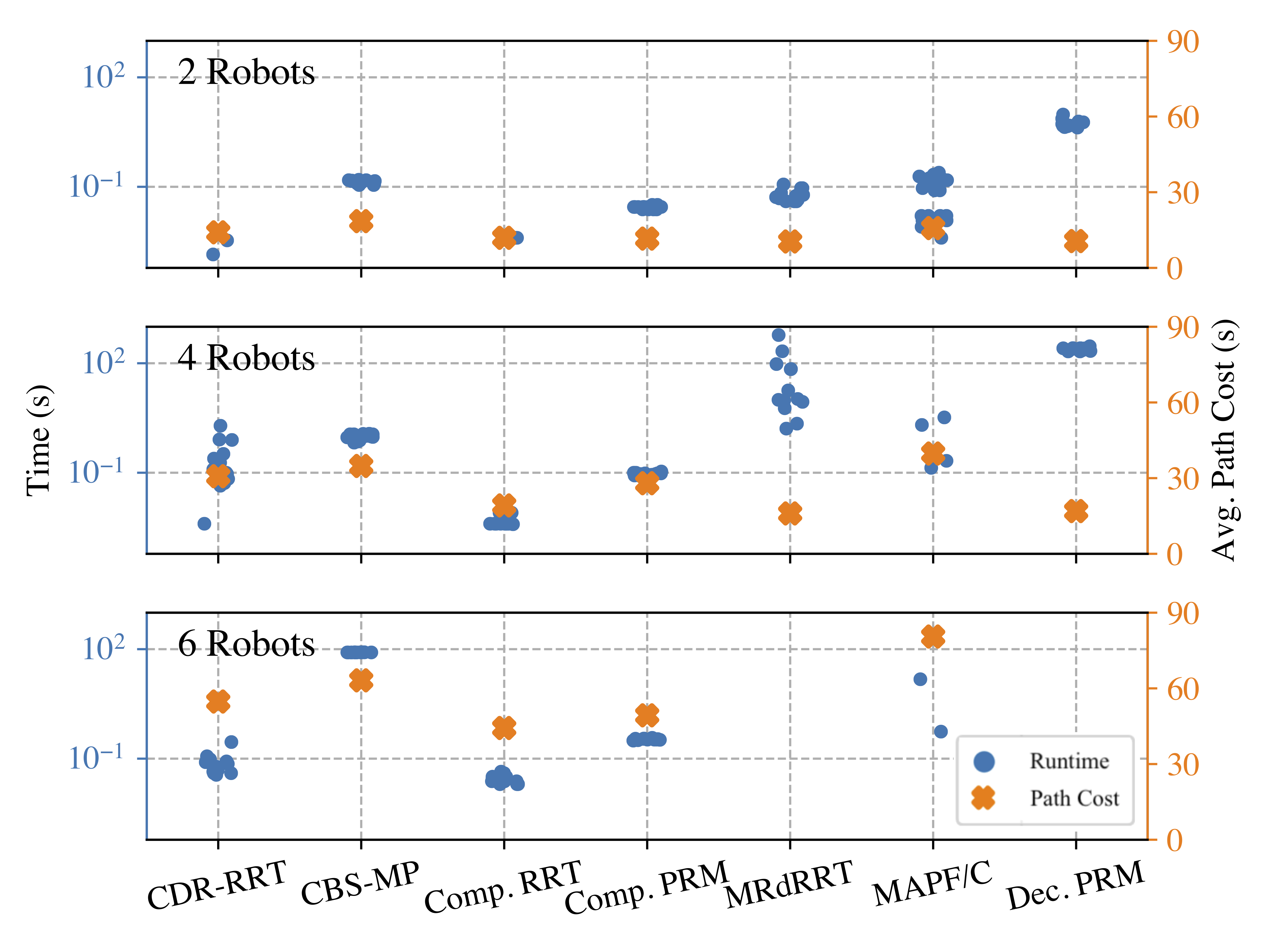}
        \label{fig:open-graph}
    }
    \subfigure[Maze Cross]{
        \includegraphics[width=.49\textwidth]{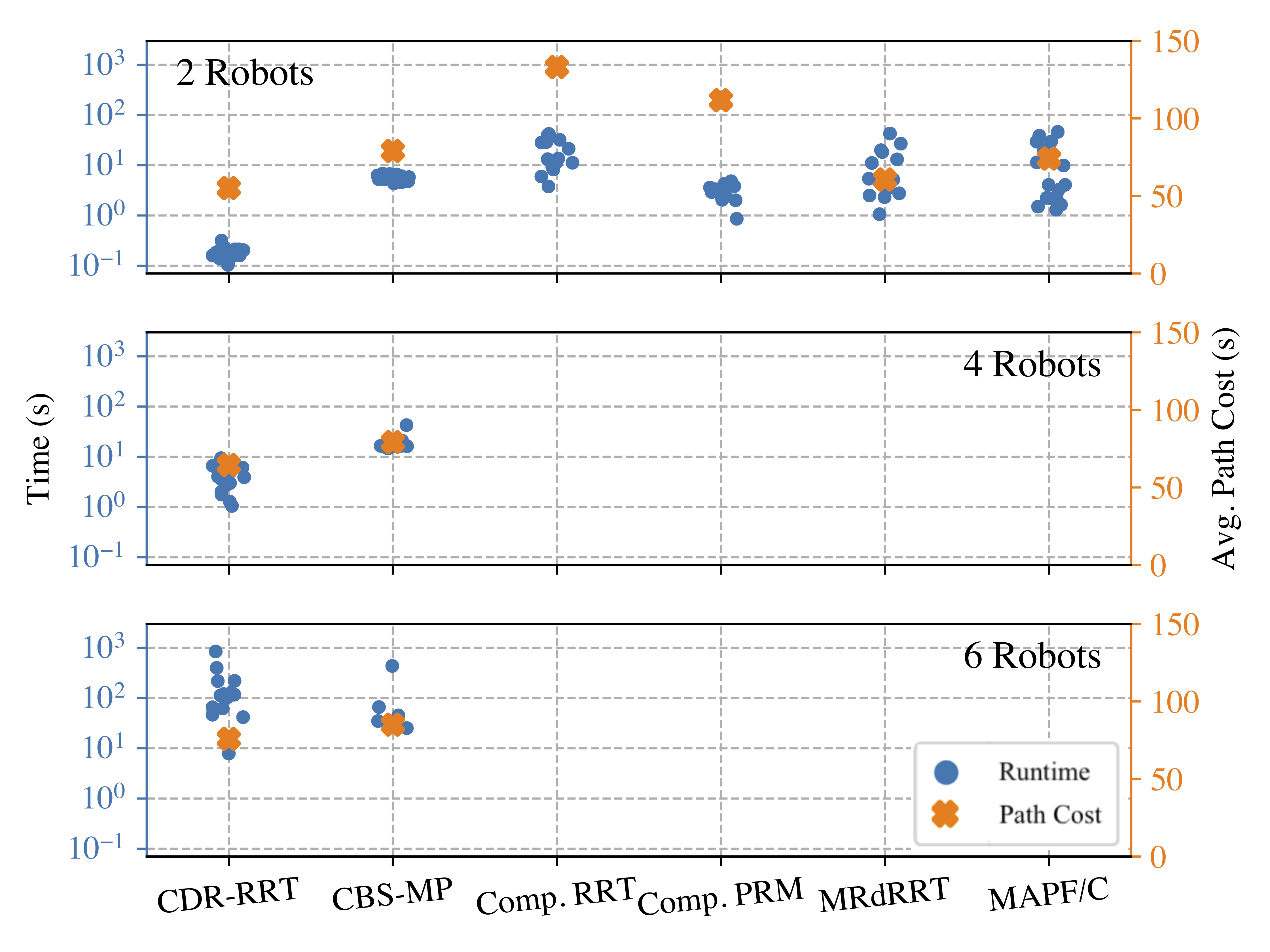}
        \label{fig:maze-graph}
    }
\caption{Results for the Open Cross (a) and Maze Cross (b) environments for all methods. Those that were unable to find a solution within the time limit are omitted. We demonstrate improved scalability as compared to the other methods by consistently achieving low planning times even as the number of robots increases. On the larger (4 and 6-robot) scenarios with narrow passages, CDR-RRT achieves the lowest running time.}
    \label{fig:cross-results}
\end{figure}

\subsection{Theoretical Analysis}
\label{sec:theory}

\red{
\textit{Theorem:} CDR-RRT is probabilistically complete.
}

\red{
\textit{Proof:}}
During each iteration of CDR-RRT, there is a probability $\epsilon > 0$ of sampling from the entire environment rather than within a region (Alg. \ref{alg:crbrrtg}, line \ref{alg:bounds}). 
Sampling from the entire environment guarantees probabilistic completeness, ensuring that a valid path from $q_{start}$ to $q_{goal}$ will be found, if one exists, even if all regions are unable to produce valid configurations.
As $\epsilon$ increases to 1 or as the size of regions is increased to encompass the entire workspace, knowledge of the workspace topology is utilized less and the method eventually reduces to Composite RRT.


\section{VALIDATION}
We run scaling MRMP queries in environments designed to highlight the strengths and weaknesses of our approach.
\red{We consider both environments with different narrow passage widths and open environments to measure how CDR-RRT compares to other state-of-the-art methods when the workspace is and is not informative. We measure each algorithm's performance in scenarios that require various levels of coordination during planning to demonstrate our improved performance when high coordination is required.}

\subsection{Experimental Setup}
We compare to several state-of-the-art MRMP methods (Table~\ref{tab:overview}).
We use CBS-MP~\cite{smsa-rmmpucs-21} with DR-PRM~\cite{suda-tgrcdrs-20} to construct the individual roadmaps as a comparison against a hybrid method with workspace guidance.
\red{We use the implementation of MAPF/C described in \cite{hpksga-tpqs-2018} with SPARS~\cite{db-spsanomp-13} roadmap generation.}
Although MRdRRT~\cite{ssh-faniaehdrfeoirimm-16} was designed primarily for manipulators, we compare to it since its use of a tensor-product roadmap to conduct an RRT-style search of the composite space is similar to our composite skeleton guidance.
\red{We pre-compute medial-axis skeletons for 2D environments and mean curvature skeletons for 3D environments.}

All methods were implemented in C++ in the Parasol Planning Library.
\red{The experiments were run in simulation with holonomic mobile robots}
using a desktop computer with an Intel Core i9-10900KF CPU at 3.7 GHz and 128 GB of RAM.
Each method is given 600 seconds to find a plan or is considered unsuccessful. 
\red{We report planning times and average path costs given by the makespan.}

\subsection{Environments}
In this section, we describe our experimental environments and explain why these scenarios highlight the advantages and disadvantages of our approach relative to other methods.

\subsubsection{\red{Corridors} (Fig.~\ref{fig:hall-cross}-\ref{fig:track})}
\red{
We consider the Hallway environment featuring a single tunnel within which only one robot may fit vertically, preventing robots from passing each other.
We consider two variants of this scenario, one in which two groups of robots start on opposite ends of the tunnel and swap places (\textit{Cross}, Fig.~\ref{fig:hall-cross}) and one in which all robots start on one side and must move to the other (\textit{Flow}, Fig.~\ref{fig:hall-flow}). In the Inlet scenario (Fig.~\ref{fig:inlet}), we show how each method performs when one robot must explicitly move out of the other's way, requiring a high level of coordination. Similarly, in the Track scenario (Fig.~\ref{fig:track}), we show how each method performs when all robots must move in the same direction around an obstacle, again requiring a high level of coordination, but for larger robot groups.}

\subsubsection{Warehouse (Fig.~\ref{fig:short-warehouse})}
The Warehouse scenario is designed to imitate the motions required to fetch or place items on shelves in a warehouse.
The topology creates several parallel narrow passages,
and queries are selected such that conflicting choices of aisles are likely, thus requiring coordination during planning to avoid inter-robot collisions.
It includes a width-wise aisle cutting through the middle of the length-wise aisles creating entry/exit points that can be used to avoid collisions.
We scale the number of aisles with the number of robots. 
We consider three variants with progressively doubled aisle widths.

\subsubsection{Open Cross (Fig.~\ref{fig:open-cross})}
We evaluate our approach on a classic open MRMP scenario~\cite{smsa-rmmpucs-21} to demonstrate how our method compares when the workspace is not informative, limiting the benefit of topological skeleton guidance.


\subsubsection{Maze Cross (Fig.~\ref{fig:3d_maze})}
In this 3D environment with a narrow maze tunnel, the number of degrees of freedom of each mobile robot is doubled relative to a 2D workspace, resulting in a very large composite space.

\subsection{Narrow Passages}


\red{
The Hallway, Inlet, and Track scenario results are given in Fig. \ref{fig:coord-graphs}.
In the Hallway scenario, CDR-RRT's use of skeleton guidance allows it to find considerably lower cost paths than other methods in both variants of the problem, especially with the larger 4 and 6-robot groups. 
Composite RRT was able to solve the 6-robot scenarios but with significantly decreased success rates and higher path costs.
MRdRRT and MAPF/C were unable to find solutions within the time limit.}

\red{Considering the Inlet scenario, runtimes for CDR-RRT, Composite RRT, Composite PRM, and MAPF/C were similar; however, the solution quality varies greatly between the methods. 
CDR-RRT has an average path cost of 15.35s, while Composite RRT, Composite PRM, and MAPF/C have average path costs of 22.31s, 22.13s, and 25.30s respectively.
By sampling along skeleton edges, CDR-RRT finds more direct, lower-cost paths.
CBS-MP and MRdRRT were unable to find solutions. 
We also demonstrate our method on physical robots (Fig.~\ref{fig:physical-inlet}). We use Turtlebot3s and integrate ROS with our planning library to follow paths generated by CDR-RRT.}

\red{In the Track environment, only CDR-RRT is able to find a solution to the 6-robot scenario due to the difficulty of sampling paths where each robot moves in the same direction. CDR-RRT's use of a MAPF heuristic to find feasible paths over the composite skeleton allows it to efficiently recognize the robots must move either clockwise or counterclockwise. None of the other methods was able to achieve over a 7\% success rate on the 4-robot scenario. Decoupled PRM was unable to find a solution for any scenario.}

The Warehouse scenarios demonstrate each method's performance in with varying widths of narrow passages (results in Fig. \ref{fig:warehouse-both-graph}).
\red{CDR-RRT achieves the lowest planning time on all scenarios and its use of skeleton guidance allows it to scale to the more complex 4 and 6-robot scenarios, where the performance of other methods significantly degrades.
Decoupled PRM was unable to complete any scenario.}

\subsection{Scalability}
\red{
We ran up to 100-robot Track scenarios with an 1800-second time limit for CDR-RRT to measure its performance on larger problems. Fig.~\ref{fig:track-cdr-rrt-graph} shows that CDR-RRT is able to efficiently find paths for very large robot teams when a high level of coordination is required. We also show that CDR-RRT's MAPF heuristic evaluation maintains scalability.
}

\subsection{Robot Crossings}

The Open Cross and Maze Cross environments evaluate each method's performance in different robot cross scenarios, in which topological guidance provides varying levels of benefit.
The Open Cross environment results are in Fig. \ref{fig:open-graph}.
When the topology is not useful, workspace guidance still biases robot paths along skeleton edges, which, in an environment without obstacles, increases the potential for collision relative to Composite RRT and Composite PRM. 
As a result, CDR-RRT has a higher average planning time than Composite RRT and Composite PRM.
This shows that composite skeleton guidance is most effective when there are narrow passages that robots must pass through.


In the Maze Cross scenarios (Fig. \ref{fig:maze-graph}),
\red{CDR-RRT and CBS-MP are the only methods able to plan for 4 robots.
Only CDR-RRT is able to plan for 6 robots 
with 100\% success within the time limit (CBS-MP - 33\%). Decoupled PRM failed to solve the 2-robot scenario.}
In 3D environments, the size of the composite \cspace{} increases significantly, 
\red{boosting the impact of composite skeleton guidance.}


\section{CONCLUSION AND FUTURE WORK}

We present Composite Dynamic Region-biased Rapidly-exploring Random Trees, a scalable workspace-guided multi-robot motion planning approach. We validate our method on a variety of environments, with and without narrow passages, to demonstrate its strengths and weaknesses. We show improved performance in constricted environments. 
Future work will explore the use of composite skeleton guidance for PRM-based roadmap construction, expanding its utility to multi-query scenarios, 
\red{as well as extending skeleton guidance to non-holonomic robot teams.}



\bibliographystyle{IEEEtran.bst}
\bibliography{robotics.bib}

\begin{thebibliography}{10}
\providecommand{\url}[1]{#1}
\csname url@rmstyle\endcsname
\providecommand{\newblock}{\relax}
\providecommand{\bibinfo}[2]{#2}
\providecommand\BIBentrySTDinterwordspacing{\spaceskip=0pt\relax}
\providecommand\BIBentryALTinterwordstretchfactor{4}
\providecommand\BIBentryALTinterwordspacing{\spaceskip=\fontdimen2\font plus
\BIBentryALTinterwordstretchfactor\fontdimen3\font minus
  \fontdimen4\font\relax}
\providecommand\BIBforeignlanguage[2]{{%
\expandafter\ifx\csname l@#1\endcsname\relax
\typeout{** WARNING: IEEEtran.bst: No hyphenation pattern has been}%
\typeout{** loaded for the language `#1'. Using the pattern for}%
\typeout{** the default language instead.}%
\else
\language=\csname l@#1\endcsname
\fi
#2}}

\bibitem{sl-uppccdpmrs-2002}
G.~Sanchez and J.-C. Latombe, ``Using a prm planner to compare centralized and
  decoupled planning for multi-robot systems,'' in \emph{Proc.\ {IEEE} Int.\
  Conf.\ Robot.\ Autom.\ ({ICRA})}, vol.~2, 2002, pp. 2112--2119.

\bibitem{l-rrtntpp-1998}
S.~M. Lavalle, ``Rapidly-exploring random trees: A new tool for path
  planning,'' Iowa State University, Tech. Rep., 1998.

\bibitem{smsa-rmmpucs-21}
I.~Solis, J.~Motes, R.~Sandstr{\"o}m, and N.~M. Amato, ``Representation-optimal
  multi-robot motion planning using conflict-based search,'' \emph{IEEE
  Robotics and Automation Letters}, vol.~6, no.~3, pp. 4608--4615, 2021.

\bibitem{wc-sefmpp-15}
G.~Wagner and H.~Choset, ``Subdimensional expansion for multirobot path
  planning,'' \emph{Artificial Intelligence}, vol. 219, pp. 1--24, 2015.

\bibitem{dsba-drbrrt-16}
J.~Denny, R.~Sandstr\"{o}m, A.~Bregger, and N.~M. Amato, ``Dynamic
  region-biased rapidly-exploring random trees,'' in \emph{Alg. Found. Robot.
  XII}.\hskip 1em plus 0.5em minus 0.4em\relax Springer, 2020, (WAFR `16).

\bibitem{suda-tgrcdrs-20}
R.~{Sandstrom}, D.~{Uwacu}, J.~{Denny}, and N.~M. {Amato}, ``Topology-guided
  roadmap construction with dynamic region sampling,'' \emph{IEEE Robotics and
  Automation Letters}, vol.~5, no.~4, pp. 6161--6168, 2020.

\bibitem{rsb-beaeisbmp-2014}
M.~Rickert, A.~Sieverling, and O.~Brock, ``Balancing exploration and
  exploitation in sampling-based motion planning,'' \emph{IEEE Transactions on
  Robotics}, vol.~30, no.~6, pp. 1305--1317, 2014.

\bibitem{lw-apcfpapo-79}
T.~{Lozano-P\'{e}rez} and M.~A. Wesley, ``An algorithm for planning
  collision-free paths among polyhedral obstacles,'' \emph{Communications of
  the ACM}, vol.~22, no.~10, pp. 560--570, Oct. 1979.

\bibitem{ss-otpmpiigtfctporam-83}
J.~T. Schwartz and M.~Sharir, ``On the “piano movers” problem. ii. general
  techniques for computing topological properties of real algebraic
  manifolds,'' \emph{Advances in applied Mathematics}, vol.~4, no.~3, pp.
  298--351, 1983.

\bibitem{c-crmp-88}
J.~F. Canny, \emph{The Complexity of Robot Motion Planning}.\hskip 1em plus
  0.5em minus 0.4em\relax Cambridge, MA: MIT Press, 1988.

\bibitem{kslo-prpp-96}
L.~E. Kavraki, P.~\v{S}vestka, J.~C. Latombe, and M.~H. Overmars,
  ``Probabilistic roadmaps for path planning in high-dimensional configuration
  spaces,'' \emph{{IEEE} Trans.\ Robot.\ Automat.}, vol.~12, no.~4, pp.
  566--580, Aug. 1996.

\bibitem{hlk-pfprp-06}
D.~Hsu, J.-C. Latombe, and H.~Kurniawati, ``On the probabilistic foundations of
  probabilistic roadmap planning,'' \emph{Int.\ J.\ Robot.\ Res.}, vol.~25, pp.
  627--643, July 2006.

\bibitem{rtla-obrrt-06}
S.~Rodriguez, X.~Tang, J.-M. Lien, and N.~M. Amato, ``An obstacle-based
  rapidly-exploring random tree,'' in \emph{Proc.\ {IEEE} Int.\ Conf.\ Robot.\
  Autom.\ ({ICRA})}, 2006.

\bibitem{yjsl-ddreecsd-05}
A.~Yershova, L.~Jaillet, T.~Simeon, and S.~M. Lavalle, ``Dynamic-domain {RRTs}:
  Efficient exploration by controlling the sampling domain,'' in \emph{Proc.\
  {IEEE} Int.\ Conf.\ Robot.\ Autom.\ ({ICRA})}, Apr. 2005, pp. 3856--3861.

\bibitem{ssh-faniaehdrfeoirimm-16}
K.~Solovey, O.~Salzman, and D.~Halperin, ``Finding a needle in an exponential
  haystack: Discrete rrt for exploration of implicit roadmaps in multi-robot
  motion planning,'' \emph{The International Journal of Robotics Research},
  vol.~35, no.~5, pp. 501--513, 2016.

\bibitem{hpksga-tpqs-2018}
W.~Hönig, J.~A. Preiss, T.~K.~S. Kumar, G.~S. Sukhatme, and N.~Ayanian,
  ``Trajectory planning for quadrotor swarms,'' \emph{IEEE Transactions on
  Robotics}, vol.~34, no.~4, pp. 856--869, 2018.

\bibitem{r-cbmrpp-2010}
M.~Ryan, ``Constraint-based multi-robot path planning,'' in \emph{2010 IEEE
  International Conference on Robotics and Automation}, 2010, pp. 922--928.

\bibitem{Yu2018}
\BIBentryALTinterwordspacing
J.~Yu and D.~Rus, \emph{An Effective Algorithmic Framework for Near Optimal
  Multi-robot Path Planning}.\hskip 1em plus 0.5em minus 0.4em\relax Cham:
  Springer International Publishing, 2018, pp. 495--511. [Online]. Available:
  \url{https://doi.org/10.1007/978-3-319-51532-8_30}
\BIBentrySTDinterwordspacing

\bibitem{mtpra-mlafsmp-04}
M.~Morales, L.~Tapia, R.~Pearce, S.~Rodriguez, and N.~M. Amato, ``A machine
  learning approach for feature-sensitive motion planning,'' in \emph{Alg.
  Found. Robot. VI}.\hskip 1em plus 0.5em minus 0.4em\relax Springer, 2005, pp.
  361--376, (WAFR `04).

\bibitem{kh-wisprp-04}
H.~Kurniawati and D.~Hsu, ``Workspace importance sampling for probabilistic
  roadmap planning,'' in \emph{Proc.\ {IEEE} Int.\ Conf.\ Intel.\ Rob.\ Syst.\
  ({IROS})}, vol.~2, Sept. 2004, pp. 1618--1623.

\bibitem{bo-uwignsprp-04}
J.~Berg and M.~Overmars, ``Using workspace information as a guide to
  non-uniform sampling in probabilistic roadmap planners,'' in \emph{Proc.\
  {IEEE} Int.\ Conf.\ Robot.\ Autom.\ ({ICRA})}, 2004, pp. 453--460.

\bibitem{kh-wco-06}
H.~Kurniawati and D.~Hsu, ``Workspace-based connectivity oracle - an adaptive
  sampling strategy for prm planning,'' in \emph{Alg. Found. Robot. VII}.\hskip
  1em plus 0.5em minus 0.4em\relax Springer, 2008, pp. 35--51, (WAFR `06).

\bibitem{pkv-mpdsclp-10}
E.~Plaku, L.~Kavraki, and M.~Vardi, ``Motion planning with dynamics by a
  synergistic combination of layers of planning,'' \emph{{IEEE} Trans.\
  Robot.}, vol.~26, no.~3, pp. 469--482, June 2010.

\bibitem{dsja-arbsfcrc-14}
J.~Denny, R.~Sandstr\"{o}m, N.~Julian, and N.~M. Amato, ``A region-based
  strategy for collaborative roadmap construction,'' in \emph{Alg. Found.
  Robot. XI}.\hskip 1em plus 0.5em minus 0.4em\relax Springer, 2015, pp.
  125--141, (WAFR `14).

\bibitem{blk-tcsrp-2012}
S.~Bhattacharya, M.~Likhachev, and V.~Kumar, ``Topological constraints in
  search-based robot path planning,'' \emph{Autonomous Robots}, vol.~33, no.~3,
  2012.

\bibitem{Blum_1967_6755}
H.~Blum, ``A transformation for extracting new descriptors of shape,'' in
  \emph{Models for Perception of Speech and Visual Form}, W.~Wathen-Dunn,
  Ed.\hskip 1em plus 0.5em minus 0.4em\relax Cambridge, MA: MIT Press, 1967.

\bibitem{t-mcs-12}
A.~Tagliasacchi, I.~Alhashim, M.~Olson, and H.~Zhang, ``Mean curvature
  skeletons,'' \emph{Computer Graphics Forum}, vol.~31, pp. 1735--1744, 08
  2012.

\bibitem{l-matoaps-1982}
D.~T. Lee, ``Medial axis transformation of a planar shape,'' \emph{IEEE
  Transactions on Pattern Analysis and Machine Intelligence}, vol. PAMI-4,
  no.~4, pp. 363--369, 1982.

\bibitem{uyma-hpwakg-2022}
D.~Uwacu, A.~Yammanuru, M.~Morales, and N.~M. Amato, ``Hierarchical planning
  with annotated skeleton guidance,'' \emph{IEEE Robotics and Automation
  Letters}, vol.~7, no.~4, pp. 11\,055--11\,061, 2022.

\bibitem{ssfs-cbsfomap-15}
G.~Sharon, R.~Stern, A.~Felner, and N.~R. Sturtevant, ``Conflict-based search
  for optimal multi-agent pathfinding,'' \emph{Artificial Intelligence}, vol.
  219, pp. 40--66, 2015.

\bibitem{ma2019searching}
\BIBentryALTinterwordspacing
H.~Ma, D.~Harabor, P.~J. Stuckey, J.~Li, and S.~Koenig, ``Searching with
  consistent prioritization for multi-agent path finding,'' in \emph{The
  Thirty-Third AAAI Conference on Artificial Intelligence}, ser. AAAI'19.\hskip
  1em plus 0.5em minus 0.4em\relax AAAI Press, 2019. [Online]. Available:
  \url{https://doi.org/10.1609/aaai.v33i01.33017643}
\BIBentrySTDinterwordspacing

\bibitem{db-spsanomp-13}
A.~Dobson and K.~E. Bekris, ``Sparse roadmap spanners for asymptotically
  near-optimal motion planning,'' \emph{Int.\ J.\ Robot.\ Res.}, 2013.

\end{thebibliography}

\end{document}